\documentclass[10pt,twocolumn,letterpaper]{article}

\usepackage[pagenumbers]{config_template/cvpr} 

\usepackage[T1]{fontenc}

\usepackage{graphicx}
\usepackage{amsmath}
\usepackage{amssymb}
\usepackage{booktabs}

\usepackage[accsupp]{axessibility}

\usepackage[pagebackref,breaklinks,colorlinks]{hyperref}

\usepackage[inline]{enumitem}

\usepackage{epsfig}
\usepackage{tabularx}

\usepackage[normalem]{ulem}

\usepackage{wrapfig}
\usepackage{subcaption}

\usepackage{url}

\usepackage{bm}
\usepackage{xspace}
\usepackage{caption}

\usepackage{balance}

\usepackage{array, makecell}

\usepackage{pifont}
\newcommand{\cmark}{\color{green}\ding{51}}
\newcommand{\xmark}{\color{red}\ding{55}}

\usepackage{multirow}

\usepackage[capitalize]{cleveref}
\crefname{section}{Sec.}{Secs.}
\Crefname{section}{Section}{Sections}
\Crefname{table}{Table}{Tables}
\crefname{table}{Tab.}{Tabs.}
\crefname{figure}{Fig.}{Figs.}
\Crefname{figure}{Figure}{Figures}

\newcommand{\vspaceTABaboveCaption}[0]{-0.50em}

\newcommand{\myarraystretch}[0]{1.1}
\newcommand{\mycolsep}[0]{4pt}

\newcommand{\mytablesetup}[0]{\renewcommand{\arraystretch}{\myarraystretch}\setlength\tabcolsep{\mycolsep}}

\newcommand{\thetaflipleft}{r(\theta_t^{l})}
\newcommand{\thetaright}{\theta_t^{r}}

\newcommand{\thetahand}{\theta_t^h}
\newcommand{\thetarefhand}{\theta^{h}_{\text{\emph{ref}}}}
\newcommand{\thetarefhandt}{\theta^{h}_{\text{\emph{ref,}}t}}

\newcommand{\thetarefhandone}{\theta_{\text{\emph{ref}},i}^{h}}
\newcommand{\thetarefhandtwo}{\theta_{\text{\emph{ref}},f}^{h}}
\newcommand{\thetarefrightone}{\theta_{\text{\emph{ref}},i}^{r}}

\newcommand{\thetarefleftone}{\theta_{\text{\emph{ref}},i}^{l}}

\newcommand{\thetarefrighttwo}{\theta_{\text{\emph{ref}},f}^{r}}

\newcommand{\thetareflefttwo}{\theta_{\text{\emph{ref}},f}^{l}}

\newcommand{\nameCOLOR}{black}

\newcommand{\sgnify}{\mbox{\textcolor{\nameCOLOR}{SGNify}}\xspace}
\newcommand{\sgnifyx}{\mbox{\textcolor{\nameCOLOR}{SGNify}}\xspace}

\newcommand{\smplifyxstar}{\mbox{SMPLify-SL}\xspace}

\newcommand{\hamnosys}{HamNoSys\xspace}

\newcommand{\pymafx}{\mbox{PyMAF-X}\xspace}

\newcommand{\spectre}{\mbox{SPECTRE}\xspace}
\newcommand{\frankmocap}{\mbox{FrankMocap}\xspace}

\newcommand{\websiteURL}{\mbox{\href{https://sgnify.is.tue.mpg.de}{sgnify.is.tue.mpg.de}}}

\newcommand{\ourTitle}{Reconstructing Signing Avatars From Video Using Linguistic Priors}

\newcommand{\fps}{\mbox{fps}\xspace}

\newcommand{\smplx}{\mbox{SMPL-X}\xspace}
\newcommand{\smplX}{\smplx}
\newcommand{\adammodel}[0]{\mbox{Adam}\xspace}

\newcommand{\smplifyx}{\mbox{SMPLify-X}\xspace}
\newcommand{\smplifyX}{\smplifyx}
\newcommand{\pixie}{\mbox{PIXIE}\xspace}

\newcommand{\ghum}{\mbox{GHUM}\xspace}

\newcommand{\openpose}{\mbox{OpenPose}\xspace}

\newcommand{\mocap}{\mbox{mocap}\xspace}

\newcommand{\twoD}{2D\xspace}
\newcommand{\threeD}{3D\xspace}
\newcommand{\fourD}{4D\xspace}

\newcommand{\slr}{\mbox{SLR}\xspace}

\newcommand{\slp}{\mbox{SLP}\xspace}

\newcommand{\slt}{\mbox{SLT}\xspace}

\newcommand{\slc}{\mbox{SLC}\xspace}
\newcommand{\slclong}{Sign Language Capture\xspace}
\newcommand{\cslc}{\mbox{CSLC}\xspace}
\newcommand{\cslclong}{Continuous Sign Language Capture\xspace}

\newcommand{\asl}{\mbox{ASL}\xspace}
\newcommand{\asllong}{American Sign Language\xspace}

\newcommand{\supmat}{\mbox{{Appx.}}\xspace}

\newcommand{\vicon}{\mbox{Vicon}\xspace}

\newcommand{\etal}{\emph{et al}.\xspace}
\newcommand{\ie}{\xspace, \emph{i.e.},\xspace}
\newcommand{\eg}{\xspace, \emph{e.g.},\xspace}

\newcommand{\rgb}{\mbox{RGB}\xspace}

\newcommand{\qheading}[1]{\noindent\textbf{#1:}}

\newcommand{\vtov}{\mbox{V2V}\xspace}
\newcommand{\TRvtov}{\mbox{TR-V2V}\xspace}

\newcommand{\upperbodylabel}{Upper Body}

\usepackage[dvipsnames]{xcolor}

\usepackage{framed}
\usepackage{flushend}
\usepackage[nounderscore]{syntax}

\begin{document}

\title{\ourTitle} 

\author{Maria-Paola~Forte \quad
Peter~Kulits \quad
Chun-Hao~Huang\quad
Vasileios~Choutas \quad
Dimitrios~Tzionas \and
Katherine~J.~Kuchenbecker \quad
Michael~J.~Black\\
{\small Max Planck Institute for Intelligent Systems, Stuttgart and T{\"u}bingen, Germany}\\
\tt\small{\{forte,kjk\}@is.mpg.de} \quad \{kulits,chuang2,vchoutas,dtzionas,black\}@tue.mpg.de
\vspace{-2.0em}
}

\twocolumn[{
    \renewcommand\twocolumn[1][]{#1}
    \maketitle
    \centering
    \vspace{+0.3em}
    \begin{minipage}{1.00\textwidth}
    \centering
    \includegraphics[width=\linewidth]{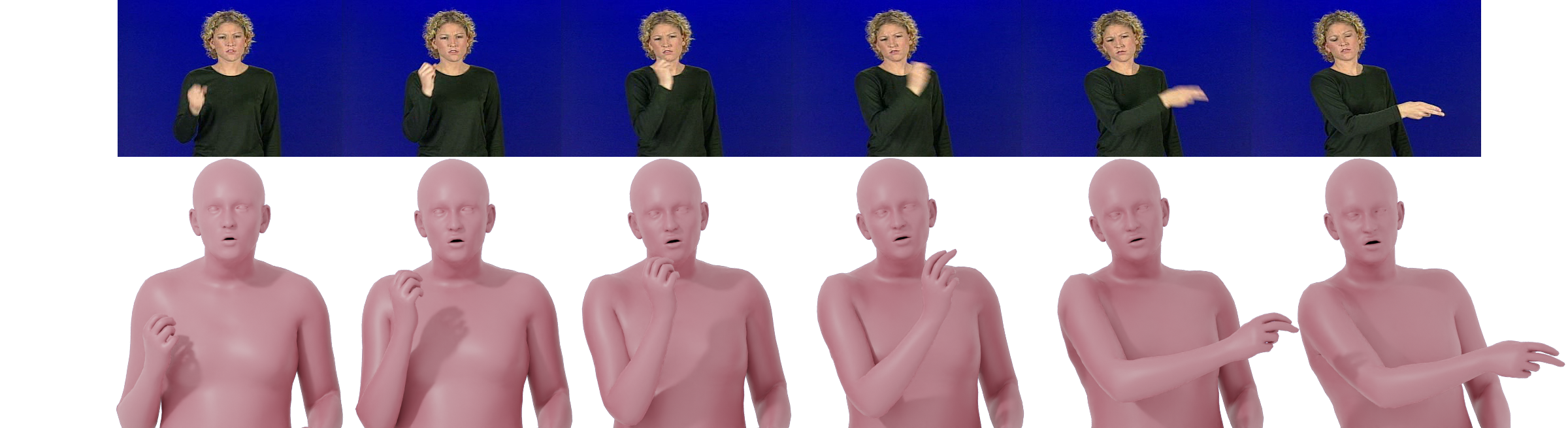}
    \end{minipage}
    \vspace{-0.1in}
    \captionof{figure}{Given a monocular, in-the-wild video of a sign-language sign, \sgnifyx automatically reconstructs a \threeD body with accurate hand pose, facial motion, and body pose. Note that motion blur obscures the finger articulations in several video frames; this is a common problem.
    Our novel linguistic priors enable accurate \threeD reconstruction despite such image degradation.}
    \label{fig:teaser}
    \vspace*{+02.50em}
    }]

\maketitle
\begin{abstract}
Sign language (SL) is the primary method of communication for the 70 million Deaf people around the world.
Video dictionaries of isolated signs are a core SL learning tool.
Replacing these with \threeD avatars can aid learning and 
enable AR/VR applications, improving access to technology and online media.
However, little work has attempted to estimate expressive \threeD avatars from SL video; occlusion, noise, and motion blur make this task difficult.
We address this by introducing novel linguistic priors that are universally applicable to SL and provide constraints on \threeD hand pose that help resolve ambiguities within isolated signs.
Our method, \sgnify, captures fine-grained hand pose, facial expression, and body movement fully automatically from in-the-wild monocular SL videos.
We evaluate \sgnify quantitatively by using a commercial motion-capture system to compute \threeD avatars synchronized with monocular video.
\sgnify outperforms state-of-the-art \threeD body-pose- and shape-estimation methods on SL videos.
A perceptual study shows that \sgnify's \threeD reconstructions are significantly more comprehensible and natural than those of previous methods and are on par with the source videos.
Code and data are available at \websiteURL.
\end{abstract}
\section{Introduction}

It is estimated that over 466 million people have disabling hearing loss~\cite{davis2019hearing} and more than 70 million people use sign language (SL) as their primary means of communication~\cite{WFD}.
Increasing use of digital communication motivates research on capturing, understanding, modeling, and synthesizing expressive \threeD SL avatars. 
Existing datasets and dictionaries used in SL recognition (\slr), translation (\slt),
and production (\slp) are primarily limited to \twoD video because the technology required to capture \threeD movement is prohibitively expensive, requires expertise to operate, and may limit the movements of the signer.
Dictionaries of isolated signs are a core SL learning tool, and many SLs have online \twoD video dictionaries.
The Deaf community is actively seeking \threeD dictionaries of isolated signs to aid learning~\cite{naert2020survey}.
The current approach to creating such \threeD signing dictionaries is fully manual, requiring an artist or a HamNoSys~\cite{hanke2004hamnosys} expert, and the resulting avatars often move unnaturally~\cite{aliwy2021development}.
We aim to automatically reconstruct expressive \threeD signing avatars from monocular SL video, which we term {\em \slclong (\slc)}.
We focus on \slc of isolated signs.

\threeD reconstruction of human pose and shape has received significant attention, but accurate \threeD hand-pose estimation remains challenging from in-the-wild video.
Challenges include the high number of degrees of freedom present in hands~\cite{bilal2011vision}, frequent occurrence of self-contact and self-occlusions~\cite{2020_interhand2.6m,Smith_elasticity_hands_2020}, low resolution, and motion blur caused by fast motions~\cite{vazquez2021isolated} that cause hand pose to be unrecognizable in many frames (see \cref{fig:teaser}).
To address these issues, we exploit the linguistic nature of SL itself to develop novel priors that help disambiguate hand poses in SL videos, leading to accurate \threeD reconstructions.
This is a novel use of linguistic ``side information'' to improve \threeD reconstruction.

Based on hand movements and poses, Battison~\cite{battison1978lexical} defines five linguistic classes that contain all SL signs.
We build on that work to define eight classes and formalize these as mathematical priors on \threeD hand shape.
We combine Battison's first two classes and place all one-handed signs in class 0, while two-handed signs are arranged in classes 1, 2, or 3, depending on how the non-dominant hand participates in the articulation of the sign. 
We then divide each of these four classes into two subclasses depending on whether the pose of the active hand(s) changes during the articulation of the sign.
We introduce two class-dependent SL linguistic constraints that capture 1) symmetry and 2) hand-pose invariance. 
Under Battison's SL symmetry condition~\cite{battison1978lexical}, when both hands actively move, the articulation of the fingers must be identical; the same is true for one class of two-handed signs in which only the dominant hand moves.
We formalize this concept as a regularization term that encourages the pose of the two hands to be similar for such signs.
Coupling the hand poses in this way effectively increases the image evidence for a pose, which improves estimates for challenging videos.
Our invariance constraint uses the observation that hand pose is either static or transitions smoothly from one pose to another during the articulation of the sign; other significant changes to hand pose are not common in SL.
Specifically, we extract a characteristic ``reference pose sequence'' (RPS) to describe each local hand pose during the sign articulation, and we penalize differences between the RPS and the estimated hand pose in each frame.
These two priors of symmetry and hand-pose invariance are universally applicable to all sign languages.

The hands alone, however, are not sufficient to accurately reproduce SL.
Information is conveyed holistically in SL through hand gestures, facial expressions, and upper-body movements in \threeD space.
To combine these, we use a \threeD whole-body model, \smplx~\cite{SMPL-X:2019}, that jointly models this information (see \cref{fig:teaser}).

Our novel hand-pose constraints are formulated to be incorporated into the loss function for training a neural network regressor or into the objective function of optimization-based methods.
In general, optimization-based methods are more computationally intensive but produce more accurate results when limited training data is available, so we take this approach here and build on the \smplifyx method~\cite{SMPL-X:2019}.
To appropriately incorporate our terms into the objective function, we need to know the class of the sign.
We train a simple model that extracts features from the raw video and determines the class to which the depicted sign belongs.
While \smplifyx is a good foundation for the hands and body, we find that it does not capture expressive facial motions well.
Consequently, we use a more expressive face regressor, \spectre~\cite{filntisis2022visual}, to capture the face parameters.
We call our method \sgnifyx.

To quantitatively evaluate \sgnifyx, we capture a native German (DGS) signer with a frontal \rgb camera synchronized with a 54-camera \vicon motion capture system and recover ground-truth meshes from the \vicon markers~\cite{AMASS_2019}.
We run \sgnifyx on the \rgb video and compute 3D vertex-to-vertex (\vtov) error between our resulting avatars and the ground-truth meshes.
We find that \sgnifyx reconstructs \smplx meshes more accurately than the competition.

We conduct a perceptual evaluation in which we present proficient signers with a video of either an estimated \smplx avatar or the real-person source video and task them with identifying the sign being performed.
Participants also rate their ease in recognizing the sign and the naturalness of the articulation.
Our results show that \sgnifyx reconstructs \threeD signs that are as recognizable as the original videos and consistently more recognizable, easier to understand, and more natural than the existing state of the art.
We also evaluate \sgnifyx in a multi-view setting and on continuous signing videos.
Despite not being designed for the latter, \sgnifyx captures the meaning in continuous SL.

\sgnifyx represents a step towards the automatic reconstruction of natural \threeD avatars from sign-language videos.
Our key contribution is the introduction of novel linguistic priors that are universal and helpful to constrain the problem of hand-pose estimation from SL video.
\sgnify is designed to work on video from different SL dictionaries across languages, backgrounds, people, trimming, image resolution, and framing, as visible in \supmat and in the video on our project page.
This capability is critical to capture \threeD signing at scale, which will enable progress on learning SL avatars.
Our code and motion-capture dataset are available for research purposes at \websiteURL.

\section{Related Work}
\qheading{Expressive \threeD Humans From \rgb Images}
Until recently, human-pose estimation has focused on the estimation of \twoD~\cite{dang2019deep} or \threeD~\cite{wang2021deep} joints of the hands and body, as well as those of facial features~\cite{bulat2017far} from single images.
In addition to methods that estimate a sparse set of landmarks, there are multiple methods that estimate the parameters of morphable models for the hand~\cite{hasson_2019_cvpr,moon2020i2l,Zimmermann_2019_ICCV,Kulon_2020_CVPR}, face~\cite{3DMM_survey,feng2021learning,guo2020towards,danvevcek2022emoca}, and body~\cite{pymaf2021,kanazawa2018end,Kolotouros_2019_ICCV,muller2021self,li2021hybrik,Kocabas_2021_ICCV,choi2020beyond,joo2020eft}.
The advent of expressive \threeD body models like \smplX~\cite{SMPL-X:2019}, \adammodel~\cite{Joo2018_adam}, and \ghum~\cite{Xu_2020_CVPR} has enabled research on estimating the full \threeD body surface~\cite{SMPL-X:2019,rong2021frankmocap,choutas2020monocular,PIXIE:3DV:2021,Xiang_2019_CVPR,Zanfir_2021_CVPR,tripathi2023ipman}.
Such body models are ideal for representing the expressiveness of SL but have rarely been applied to this domain~\cite{kratimenos2021independent}.

\qheading{Human Pose for Sign Language}
To enable detailed \threeD pose estimation from images, How2Sign~\cite{duarte2021how2sign} provides \threeD skeleton reconstructions for three hours of data captured in a Panoptic Studio~\cite{Joo_2015_ICCV}. 
However, the skeletal representation lacks the richness of a full \threeD body model and omits surface details that are important for communication~\cite{moryossef2021evaluating}.
Kratimenos \etal~\cite{kratimenos2021independent} use \smplifyx to estimate \threeD pose and shape on the GSLL sign-language dataset~\cite{theodorakis2014dynamic}.
They compare SL recognition accuracy using features from raw \rgb images,
\openpose~\cite{openpose} \twoD skeletons, and \smplX bodies and observe the best automated recognition results with \smplX, illustrating the benefit of using a \threeD model. 
They also highlight the importance of capturing the face and body; in an ablation study, they show that neglecting the face and body harms recognition accuracy~\cite{kratimenos2021independent}. 
However, their \smplX reconstructions use existing off-the-shelf methods and lack visual realism.
\smplifyx~\cite{SMPL-X:2019} and other recent \threeD pose-reconstruction methods~\cite{rong2021frankmocap,PIXIE:3DV:2021}, as well as keypoint detectors, struggle when applied to SL video due to challenging self-occlusion, hand--hand and hand--body interactions~\cite{moryossef2021evaluating}, motion blur~\cite{vazquez2021isolated}, and cropping inherent to SL.
SignPose~\cite{krishna2021signpose} is a \threeD-pose-lifting method for SL; it uses manually created synthetic SL animations to infer a textured avatar from single \rgb images.
SignPose requires all \openpose keypoints above the pelvis to be detected, which is unrealistic in noisy SL videos.
We address these challenges by incorporating sign-language knowledge in the form of linguistic constraints.
Since the early 2000s, the integration of linguistic information has been known to be beneficial to both \slr~\cite{cooper2011sign} and \slp~\cite{marshall2003prototype}, but this strategy has not previously been applied to \slc.

\begin{figure*}[t]
\centering
\includegraphics[width=0.91\linewidth]{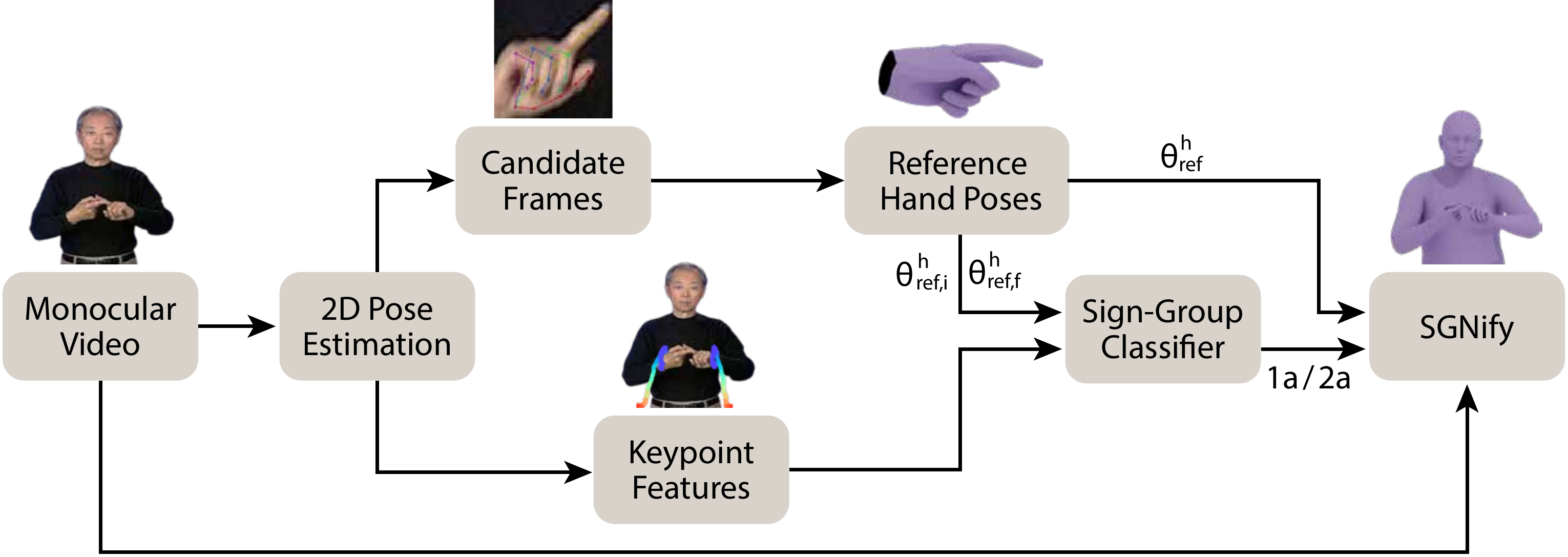}
\caption{
Given a video of a sign-language (SL) sign as input, our method preprocesses the data to first extract \twoD keypoints.
The hand keypoints are used to select candidate frames for estimating the reference hand poses ($\thetarefhandone$, $\thetarefhandtwo$, and, for static hand poses, also $\thetarefhand$).
The initial and final reference hand poses ($\thetarefhandone$ and $\thetarefhandtwo$), together with wrist-keypoint features detected across the sequence, are then fed into our sign-group classifier, which automatically classifies signs in monocular SL video into six groups based on linguistic rules universally applicable to SL~\cite{battison1978lexical}.
Using the predicted group labels and the relevant reference hand poses, \sgnify applies the appropriate linguistic constraints to improve SL \threeD hand-pose estimation, especially when the video frame is ambiguous.
}
\label{fig:model_diagram}
\end{figure*}
\section{Method}
We introduce \sgnifyx, an offline method for reconstructing \threeD body shape and pose of SL from monocular \rgb video. 
\sgnifyx centers around a key insight: SL signs follow universal linguistic rules that can be formulated as class-specific priors and used to improve hand-pose estimation.
Our full pipeline is shown in \cref{fig:model_diagram}.

\subsection{\bf \smplifyxstar: Baseline for Sign-Language Video} \label{sec:smplifyxSL}
Our baseline method builds on \smplifyX~\cite{SMPL-X:2019}, which estimates \smplX~\cite{SMPL-X:2019} parameters from \rgb images.
\smplX is a \threeD body model, representing whole-body pose and shape, including finger articulations and facial expressions.
\smplx is a function, $M(\theta, \beta, \psi)$, parameterized by body pose $\theta$ (including hand pose $\theta_h$), body shape $\beta$, and facial expressions $\psi$, that outputs a \threeD body mesh.

To create a strong baseline, we extend \smplifyX to video by adapting it
in the following ways:
\begin{enumerate*}[label=(\arabic*)]
\item We cope with the upper-body framing typical of SL videos by changing the heuristic used for camera initialization and the estimation of the out-of-view lower-body joints.
\item Since human motion is locally smooth in time, we initialize $\theta_{t}\in\mathbb{R}^{|\theta|}$ with $\theta_{t-1}$ and include a zero-velocity loss on the hands and body to encourage smooth reconstructions.
\item We estimate shape parameters ($\beta$) over multiple frames by taking the median of the parameter estimates and not-optimizing them during the per-frame reconstruction.
\item To better capture the frequent hand--hand and hand--body interactions (mainly with the face and the chest), we employ the more robust self-contact loss of M{\"u}ller \etal~\cite{muller2021self} instead of the original \smplifyX interpenetration term.
\item For each frame, we pre-compute the facial expressions ($\psi$) and jaw poses with \spectre~\cite{filntisis2022visual}.
These parameters are substituted into \smplx at the end of the optimization.
\spectre can be swapped for any method whose expression parameters are consistent with those of \smplx\eg EMOCA~\cite{danvevcek2022emoca}.
We denote the baseline \smplifyxstar.
\end{enumerate*}

\subsection{Linguistic Constraints}
\label{sec:lingconstr}

State-of-the-art optimization- and regression-based human pose estimation methods struggle on SL video, particularly with the estimation of hand pose.
We address this challenge by formulating linguistic constraints as additional losses on hand pose and integrating them into the \smplifyxstar objective function.
First, we adapt the five sign-classification and morpheme-structure conditions introduced for \asllong (\asl) by Battison~\cite{battison1978lexical} to divide signs into four primary classes:

\qheading{Class 0}
one-handed signs in which only the dominant hand articulates the sign.

\qheading{Class 1}
two-handed signs in which both hands are active.
They share the same poses and perform the same movement in a synchronous or alternating pattern. This class includes all signs that follow Battison's symmetry condition~\cite{battison1978lexical}.

\qheading{Class 2}
two-handed signs in which the dominant hand is active, the non-dominant hand is passive (its position and pose do not change during the articulation of the sign), and the two hands have the same initial pose.

\qheading{Class 3}
two-handed signs in which the dominant hand is active, the non-dominant hand is passive, and the two hands have different hand poses.
All signs in this class follow Battison's dominance condition~\cite{battison1978lexical}.

We further divide each class into two subclasses: \emph{subclass a} contains signs in which the hand pose of the active hand(s) does not change throughout the articulation of the sign (\textbf{static}), and \emph{subclass b} contains all signs in which the hand pose changes (\textbf{transitioning}). 

Note that the division into these classes is not limited to \asl; Eccarius \etal~\cite{eccarius2007symmetry} show that the phonological and prosodic properties of \asl can be successfully transferred to other sign-language lexicons.

We then convert these linguistic classes into two \threeD pose constraints: hand-pose symmetry and hand-pose invariance.
Signs in the same class share the same constraints (see \cref{tab:classes} and \cref{tab:classes_images} in \supmat).

Below we describe only the new terms added to the \smplifyx objective.
Please see \supmat for the full \sgnifyx objective.

\begin{table}[!t]

\centering
\mytablesetup
\vspace{\vspaceTABaboveCaption}
\begin{center}
\begin{tabular}{cccc}
\toprule
\multirow{2}{*}{~~Class~~} & \multirow{2}{*}{\begin{tabular}[c]{@{}c@{}}~~Hand-Pose~~\\~Symmetry\end{tabular}} &
\multicolumn{2}{c}{Hand-Pose Invariance} \\ 
&  & Dominant & Non-dominant \\
\midrule

{0a} & \xmark & static & \xmark \\
0b & \xmark & transitioning & \xmark \\
1a & \cmark & static & static \\
1b & \cmark & transitioning & transitioning \\
2a & \cmark & static & static \\
2b & \xmark & transitioning & static \\
3a & \xmark & static & static \\
3b & \xmark & transitioning & static \\
\bottomrule
\end{tabular}
\end{center}
\vspace{-0.15in}
\caption{Linguistic constraints defining the eight sign classes.}
\label{tab:classes}
\vspace{-00.50em}
\end{table}
\subsubsection{Hand-Pose Symmetry}

We encourage the left and right hand poses to match for the relevant classes (classes 1a, 1b, and 2a in \cref{tab:classes}):
\begin{align} \label{eq:symmetry}
  L_s = \lambda_s||\thetaright - \thetaflipleft||^2_2,
\end{align}
where $\thetaright$ is the finger articulation of the right hand, and $\thetaflipleft$ is a reflection function to represent the articulation of the fingers of the left hand as if it were a right hand.
This loss penalizes differences in finger poses between the hands.\looseness=-1
\subsubsection{Hand-Pose Invariance}
Each sign has a characteristic reference hand pose sequence (RPS).
The RPS defines the hand pose that we expect at each time $t$ during the articulation of the sign. 
The hand-pose-invariance constraint penalizes differences between the reference hand pose $\thetarefhandt\in\textrm{RPS}^{h}$ and the estimated hand pose $\thetahand$: 
\begin{align} \label{eq:invariance}
L^{h}_{i} =  \lambda_{i}||\thetarefhandt - \thetahand||^2_2 ,
\end{align}
where $h$ represents either the left or the right hand. 

Throughout each sign, the hand pose either stays static or transitions between two poses.
When static, only one hand pose, $\thetarefhand$, is representative of the RPS.
Signs where the hand pose is transitioning are characterized by two reference hand poses, $\thetarefhandone$ and $\thetarefhandtwo$, corresponding respectively to the initial and final poses.
We interpolate $\thetarefhandone$ and $\thetarefhandtwo$ with spherical linear interpolation~\cite{shoemake1985animating} to obtain intermediate poses.
We presently do not consider signs with repeated hand-pose transitions\eg STORY in \asl, which occur in a small percentage of signs ($\sim$3\%).

\subsection{Automatization}
To work fully automatically, \sgnifyx must 1) estimate the poses needed to enforce the hand-pose-invariance constraint and 2) classify which sign group is present in a video sequence (see \cref{fig:model_diagram}).

To estimate the reference hand poses ($\thetarefhand$, $\thetarefhandone$, and $\thetarefhandtwo$), our method selects candidate frames in the core part of the sign using hand-keypoint detection confidences, and it uses \smplifyx (adapted to SL cropping) to reconstruct a preliminary \threeD hand pose for each candidate frame.
With static hand poses, $\thetarefhand$ is obtained by taking the average hand poses of these candidates.
With transitioning hand poses, the core part of a sign is divided into two intervals, and $\thetarefhandone$ and $\thetarefhandtwo$ correspond to the average hand poses of the candidate frames in the first and second intervals, respectively (see \supmat for more details).

The constraints applied to each sign depend on its sign group; we have six sign groups because classes 1a \& 2a share the same constraints, as do 2b \& 3b (see \cref{tab:classes}).
There is insufficient paired data to train a CNN classifier, so 
we use an intuitive and interpretable decision tree trained on extracted \twoD and \threeD pose features.
\iftrue{Our features are invariant to the handedness of the signer and include:
\begin{enumerate*}[label={\arabic*})]
    \item the minimum of the maximum height differences of each wrist across the sequence:
    $\min(\{w_r\}_{\max}-\{w_r\}_{\min},\{w_l\}_{\max}-\{w_l\}_{\min})$, where $w_l$ and $w_r$ are the heights of the left and right wrists, respectively.
     \item the cosine distance between the initial poses of each hand:
      $\text{CosDist}(\thetarefrightone, \thetarefleftone)$,
     \item the maximum of the two cosine distances between each initial and final hand pose:
    $\max(\text{CosDist}(\thetarefrightone, \thetarefrighttwo), \text{CosDist}(\thetarefleftone, \thetareflefttwo))$.
\end{enumerate*}
}
\else{
Our features include:
\begin{itemize}
    \item the minimum of the maximum height differences of each wrist across the sequence
    \begin{align}
        \min(\{h_r\}_{\max}-\{h_r\}_{\min},\{h_l\}_{\max}-\{h_l\}_{\min}),
    \end{align}
    \item the cosine distance between the initial RPS poses of each hand
     \begin{align}
     \text{CosDist}(\thetarefrightone, \thetarefleftone),
    \end{align}
    \item the maximum of the two cosine distances between each initial and final RPS hand pose,
    \begin{align}
        \max(\text{CosDist}(\thetarefrightone, \thetarefrighttwo), \text{CosDist}(\thetarefleftone, \thetareflefttwo)).
    \end{align}
\end{itemize}
}
\fi

\begin{figure}[t]
\centering
\includegraphics[width=\linewidth]{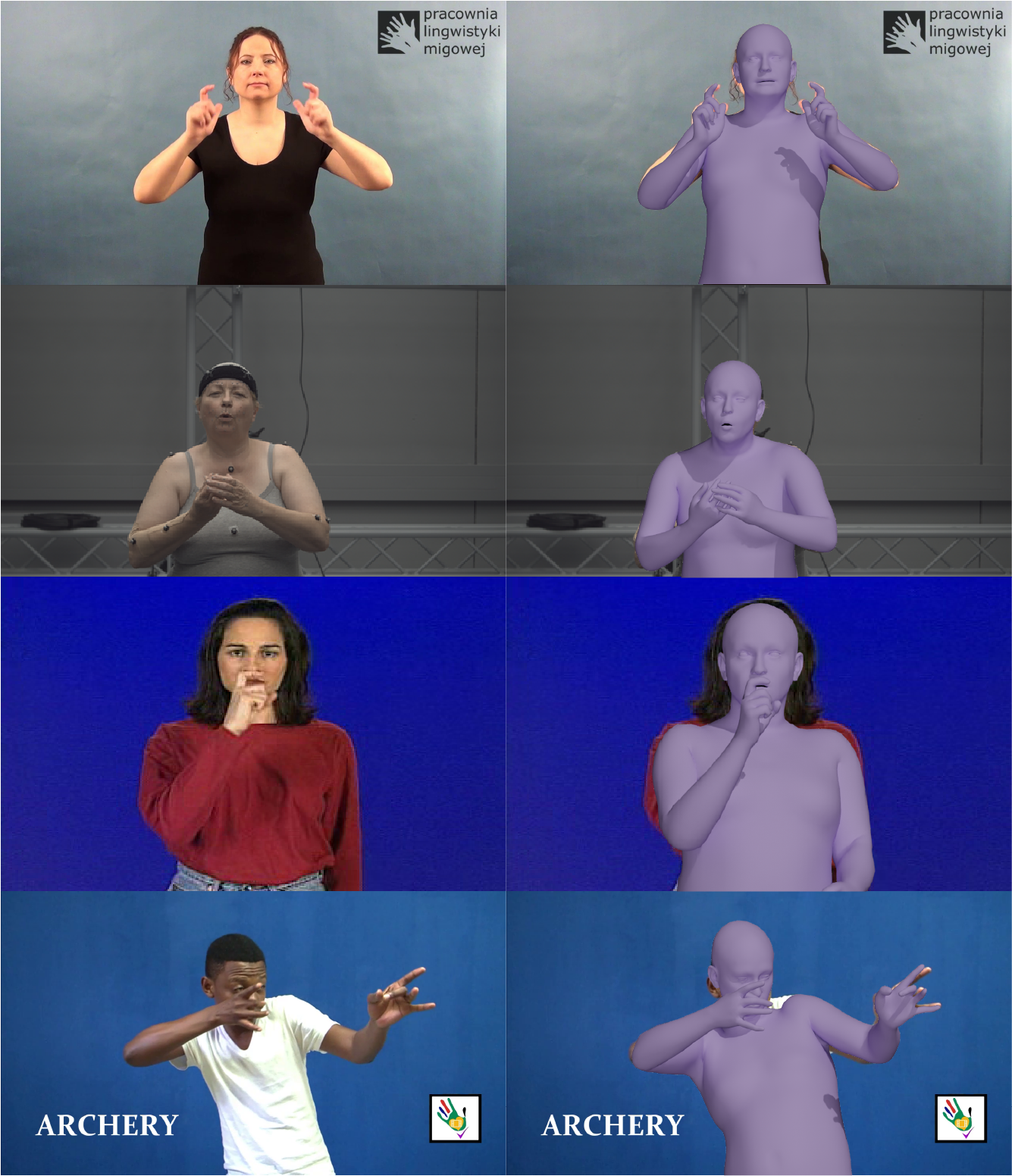}
\caption{
Samples frames reconstructed by \sgnify.
The input videos are from diverse multilingual datasets.
Row~1: Polish SL sign GDAŃSK (GDANSK, class 1a) from the dataset~\cite{slownikpjm} used for training our sign-group classifier.
Row~2: DGS sign BLUME (FLOWER, class 3b) from our captured dataset used in the quantitative evaluation.
Row~3: ASL sign DOLL (class 0a) from the dataset~\cite{tennant2010american} used in the perceptual study.
Row~4: South African SL sign ARCHERY (class 2a) from the in-the-wild dataset~\cite{reaslsasl}.
}
\label{fig:inthewild}
\end{figure}

We train our sign-group classifier on the over 3,000 videos from the Corpus-based Dictionary of Polish Sign Language (CDPSL)~\cite{slownikpjm}; 
these are annotated with \hamnosys~\cite{hanke2004hamnosys}.
We construct a grammar to convert \hamnosys annotations into our group labels (see \supmat)
Row 1 of \cref{fig:inthewild} shows a sample frame from CDPSL.
This dataset is not used in our quantitative analysis or perceptual study.

\subsection{\sgnifyx Extensions}
First, we follow Huang \etal~\cite{RICH} to extend \sgnifyx to work on multi-view videos.
Second, we propose a baseline method for continuous \slc (\cslc).
\cslc introduces additional challenges, such as the segmentation of sentences into signs; this is an active field of research. When a sentence is given as input, we use Renz \etal~\cite{Renz2021signsegmentation_b} to segment the input video and then process each segment with \sgnifyx.
The first frame of each segment is initialized from the last frame of the previous one.
These extensions are not our main contribution and are described in \supmat

\section{Dataset}
\label{sec:dataset}
To quantitatively evaluate \sgnifyx as a viable method for \slc, we collected motion-capture data with ground-truth \smplx bodies articulating signs. 
Our dataset represents the first publicly available expressive full-\fourD capture of isolated SL signs.
The experimental procedure was reviewed by the ethics council of the University of T{\"u}bingen without objections or remarks (709/2021B02).

 \begin{table}[t!]
\mytablesetup
 \centering
 \begin{tabular}{ccccccccc}
 \toprule
 \textbf{Class} & 0a & 0b & 1a & 1b & 2a & 2b & 3a & 3b   \\
 \textbf{\# Signs} & 12 & 3 & 14 & 3 & 11 & 2 & 10 & 2    \\
 \bottomrule
 \end{tabular}
 \caption{Number of signs captured for each class.}
 \label{tab:corpus}
 \vspace{-00.50em}
 \end{table}

In consultation with a Deaf DGS teacher and a DGS interpreter, we defined a German SL corpus consisting of $57$ isolated signs.
The selected signs cover a wide range of challenges for \slc, such as self-contact and self-occlusion.
\Cref{tab:corpus} summarizes the number of signs collected for each of the eight classes.
Signs of \emph{subclass b} are less common, and this is reflected in our corpus.

We captured a native right-handed DGS signer with a \vicon \mocap system at $120$ \fps, synchronized with a frontal $4112\times3008$ \rgb camera at $60$ \fps\/, framing an upper-body view as typically found in SL video.
The hands start and end at rest at the signer's sides, and each sign lasts between $1.7$ and $3.5$ seconds after trimming.
In total, our dataset comprises 16,608 \mocap frames and 8,304 \rgb frames.
To obtain ground-truth \smplx meshes, we scanned the participant in a \fourD body scanner in several poses.
The \smplx mesh was registered to these scans and averaged to obtain a personalized body-shape mesh.
MoSh++ was then used to fit this mesh to the \mocap markers~\cite{AMASS_2019}.
Marker-based \mocap is useful for evaluating ground truth but is not practical for SLC at scale:
it is expensive and requires expertise, the reflective markers attached to the signer can influence contact-heavy motions, and processing the resulting data is time consuming.
If our monocular method can approach the performance of \mocap, it will be widely applicable.
\section{Experiments}

\subsection{Quantitative Evaluation}
\label{sec:experiment}

\begin{table}[t!]
\mytablesetup
\centering
  \resizebox{0.99\linewidth}{!}{
\begin{tabular}{cccc}
\toprule
\textbf{Method} & \textbf{\upperbodylabel} & \textbf{Left Hand} & \textbf{Right Hand} 
\\
\midrule 

    \frankmocap~\cite{rong2021frankmocap} & 78.07 & 20.47 & 19.62  \\
    \pixie~\cite{PIXIE:3DV:2021}  & 60.11 & 25.02 & 22.42 \\
    \pymafx~\cite{zhang2022pymaf}  & 68.61 & 21.46 & 19.19 \\
    \smplifyxstar & 56.07 & 22.23 & 18.83 \\
    \sgnifyx & \textbf{55.63} & \textbf{19.22} & \textbf{17.50} \\

\bottomrule
\end{tabular}
}
\vspace{\vspaceTABaboveCaption}
\caption{
Evaluation on our ground-truth \mocap dataset: mean \TRvtov error (mm) for five methods and three body regions.}
\label{tab:v2v}
\end{table}
\begin{table}[t!]
    \centering
    \mytablesetup
  \resizebox{0.99\linewidth}{!}{
        \begin{tabular}{cccccc}
    \toprule
    \textbf{Method} & Sym & Inv &  \textbf{Left Hand} & \textbf{Right Hand}  & \textbf{Both Hands}   \\
    \midrule
    \smplifyxstar & \xmark & \xmark & 20.30 & 18.78 & 19.54 \\
    \sgnify & \cmark & \xmark & 18.44 &  18.39  & 18.41  \\
    \sgnify & \xmark & \cmark & 19.76 & \textbf{17.16}  & 18.46 \\
    \sgnifyx & \cmark & \cmark & \textbf{17.72} & 17.29  & \textbf{17.50} \\

    \bottomrule
\end{tabular}
    }
    \vspace{\vspaceTABaboveCaption}
    \caption{Evaluating how the linguistic constraints of symmetry and invariance affect mean \TRvtov error (mm) in symmetric signs.}
    \label{tab:v2vi1_sym}
\end{table}
\begin{table}[t!]
    \centering
    \mytablesetup

    \resizebox{1.00\linewidth}{!}{
    \begin{tabular}{cccccc}
    \toprule
    \textbf{Method} & Inv & \textbf{Left Hand} & \textbf{Right Hand} & \textbf{Both Hands}
    \\
    \midrule
    \smplifyxstar & \xmark & 26.09 & 18.89 & 20.50 \\
    \sgnifyx & \cmark & \textbf{22.22} & \textbf{17.70} & \textbf{18.60}  \\
    \bottomrule
\end{tabular}
    }
    \vspace{\vspaceTABaboveCaption}
    \caption{Evaluating how the linguistic constraint of hand-pose invariance affects mean \TRvtov error (mm) in asymmetric signs.}
    \label{tab:v2vi1_nosym}
    \vspace{-00.50em}
\end{table}

We quantitatively evaluate \sgnify, compare it with state-of-the-art methods, and quantify the improvement derived from each linguistic constraint.
To emulate in-the-wild data, which might have very low resolution, low framerate,
and an occluded lower body,
we pre-processed our high-quality video data to a resolution of $514\times300$ at $30$ \fps\/, and we cropped the input images above the pelvis (see Row 2 of \cref{fig:inthewild}).
We used the synchronized meshes captured from the observed \vicon markers~\cite{AMASS_2019} as ground truth for evaluation.
Since all tested methods estimate \smplX meshes with the same topology, we compute the mean per-vertex error (\TRvtov) by considering the vertices above the pelvis. 
The prefix ``TR'' means that we translationally align the mesh reconstructed for each frame with the ground truth; \emph{i.e.}, we center the meshes before computing these errors.
Since the starting and ending transitions are not part of the sign itself, we manually annotate the expressive central portion of each sign from the raw videos and compute the quantitative results on only these central frames (in total, 2,872 \rgb frames).

\Cref{tab:v2v} shows the mean \TRvtov error across the $57$ signs for four methods and three body regions.
The columns labeled ``\upperbodylabel'' report the error computed when considering the hands and the upper body (vertices above the pelvis).
We include the head but not the face because our \mocap system struggles to reconstruct face details using only $27$ markers.
We separately report the \TRvtov errors for the left and right hands because of the central role they play in SL.
We provide a visualization of the vertices selected for each of these evaluations in \supmat 
This experiment compares \sgnifyx with \frankmocap~\cite{rong2021frankmocap}, \pixie~\cite{PIXIE:3DV:2021}, \pymafx~\cite{zhang2022pymaf}, and our baseline \smplifyxstar. 
\sgnifyx achieves the lowest error for the upper body and both hands, beating the state-of-the-art methods.

\Cref{tab:v2vi1_sym,tab:v2vi1_nosym} show the improvements derived
from each linguistic constraint.
\Cref{tab:v2vi1_sym} reports the mean \TRvtov in symmetric signs (classes 1a, 1b, and 2a) when using no constraints\ie \smplifyxstar (Row 1), only one constraint (Rows 2 and 3), or both linguistic constraints (Row 4).
When one linguistic constraint is used, the \TRvtov of both hands decreases.
We observe that the symmetry constraint has the overall greater effect of the two.
When the non-dominant hand is passive, it is often rotated at an angle difficult to capture from a frontal camera; this behavior might explain the greater effect of the symmetry constraint on the left hand. 
When both constraints are applied, we observe a more substantial decrease in the \TRvtov error of the left hand and a slight increase in the error of the right hand compared to when only the hand-pose invariance is applied to the right hand.
We believe this happens because the symmetry constraint enforces symmetry between the two hands without knowing which hand reconstruction is more correct, and, for the same reason as above, detecting an accurate RPS for the non-dominant hand is often more challenging; overall, however, using both constraints greatly benefits the reconstruction.
In \cref{tab:v2vi1_nosym}, we separately evaluate performance on signs that do not present symmetry between the two hands (classes 0a, 0b, 2b, 3a, and 3b).
As expected, the invariance prior included in \sgnifyx improves the performance on all metrics.
These quantitative results indicate that our linguistic constraints improve the reconstructions.\looseness=-1

\subsection{Perceptual Study}

We conduct an online perceptual study to 1) compare \sgnifyx with the best-performing state-of-the-art method for SL and 2) evaluate the improvement derived from the linguistic constraints (\sgnifyx vs.\ \smplifyxstar).
Even though \frankmocap has a lower hand error than \pymafx and \pixie, it has significant errors in the lower body when the full body is not seen, and higher errors in the upper body; these errors greatly affect the perceptual experience, making it unsuitable for the SL task, as already noticed in~\cite{krishna2021signpose}.
We thus use \pymafx since its hand-pose estimates are more accurate than \pixie (see \cref{tab:v2v}). 

Approval of all experimental procedures for this study was granted by the Ethics Council of the Max Planck Society under the Haptic Intelligence Department’s framework agreement under protocol number F027A.
No participants are employed by our institution, and all are compensated 20~USD for their time.

Our study involves 20 adult participants who all stated that they have an advanced level of proficiency (expert level) in ASL.
We discarded responses from other potential participants who did not correctly recognize at least 70\% of the signs presented via real-person video.
15 of the final 20 participants (75\%) are Deaf, and one participant is left-handed.
Participant ages are 46.75 $\pm$ 11.78.

We used \sgnify, \smplifyxstar, and \pymafx to reconstruct avatars from $50$ videos taken from The American Sign Language Handshape Dictionary~\cite{tennant2010american}; see Row 3 of \cref{fig:inthewild} for an example.
After responding to demographic questions, each participant evaluates the same six training videos to calibrate their responses to the quality of the presented reconstructions.
We divide the remaining 44 signs into four batches (real-person \rgb video, \sgnify, \smplifyxstar, and \pymafx) that are balanced by sign class and sign frequency.
We include real-person video to obtain an upper-bound on recognition performance and, as mentioned, to filter participants.
We assign each participant to one of four surveys.
Each survey contains all 44 test signs, and the four methods are rotated through the four sign batches across surveys.
We further shuffle the questions in each survey for each user.

The participant enters the sign they believe the avatar or the real person is articulating in each video.
They rate their ease in recognizing the presented sign using a visual analog scale (VAS) ranging from 0 to 100 with five standard labels ranging from ``very difficult'' to ``very easy.''
They also evaluate the naturalness of the articulation on a VAS-labeled scale from ``very unnatural'' to ``very natural.''
Participants are able to replay each video.
We provide additional space for comments for each sign and at the end of the study.
The self-reported participant carefulness is 84.45 $\pm$ 11.44 on a scale from 0 to 100.

\begin{figure*}[t]
\centering
\includegraphics[width=0.95\linewidth,trim=4cm 0 0 1cm,clip]{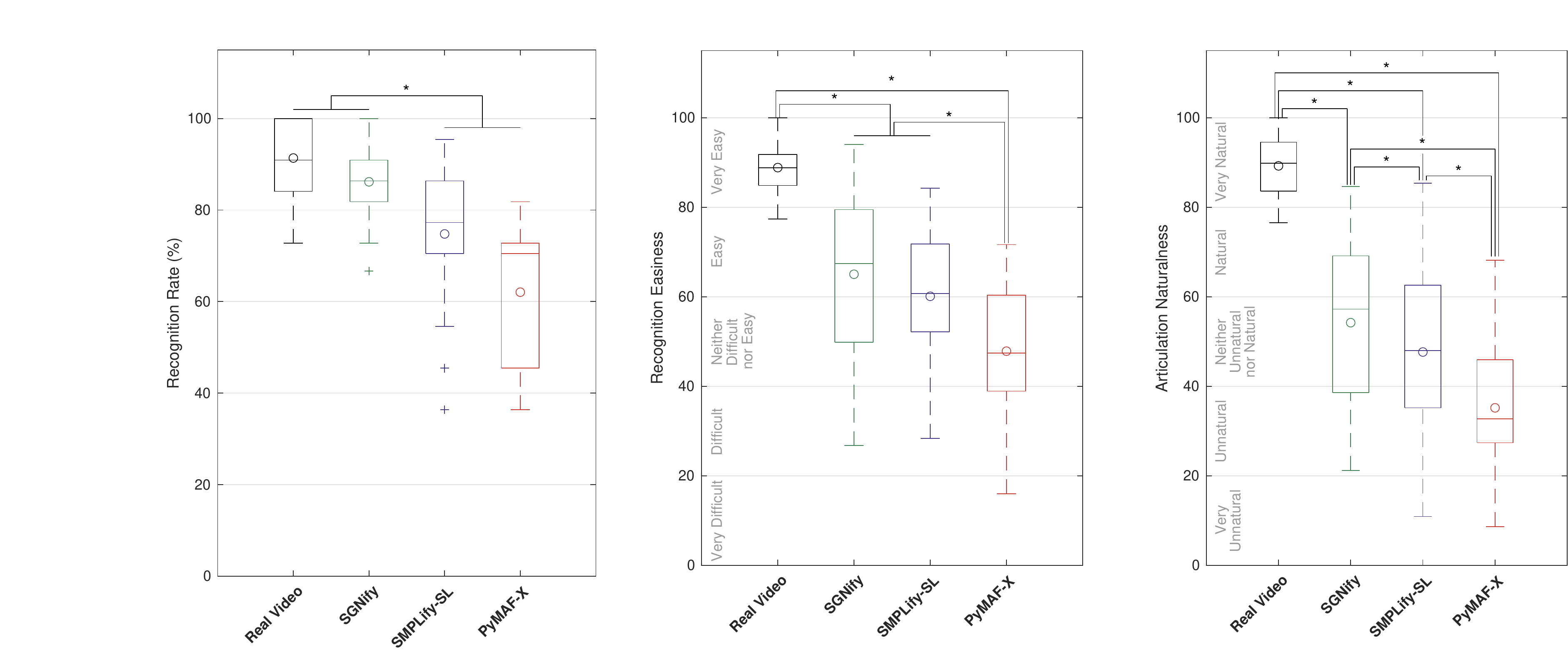}
\caption{Box plots of the data from the perceptual study.
The circle shows the mean, and the central line shows the median. The top and bottom box edges show the interquartile range (IQR), and the whiskers encompass the range up to 1.5 times the IQR.
Outliers are marked with +. Statistically significant pairwise differences are indicated with a line and a $\star$.
Left: Distribution of the rate at which participants recognized signs presented with each of the four methods.
\mbox{\emph{Real Video}} and \emph{\sgnify} achieve significantly higher recognition rates than both \emph{\smplifyxstar} and \emph{\pymafx}.
Center: Distribution of the average easiness ratings participants assigned to recognizing signs presented by the four methods.
All pairwise combinations except \mbox{\emph{\sgnify--\smplifyxstar}} are significantly different.
Right: Distribution of the average articulation naturalness ratings participants assigned to the four methods.
All pairwise combinations are significantly different.}
\label{fig:percres}
\end{figure*}

The sign annotations submitted by participants are graded as either ``incorrect'' (no credit), ``partially correct'' (half credit), or ``completely correct'' (full credit).
We calculate the rates at which each participant recognized the signs for each method.
We visualize the resulting 20 $\times$ 4 matrix of recognition rates with box plots in the left plot of \cref{fig:percres}.
Participants recognize signs in real-person videos with an average accuracy of 90.9\% and signs reconstructed by \sgnify with 86.2\% accuracy.
Signs reconstructed with \smplifyxstar and \pymafx are recognized less accurately, at 74.8\% and 62.0\%, respectively.
We evaluate the statistical significance of these recognition rates. 
Some distributions failed a Shapiro-Wilk normality test, so we used the non-parametric Friedman test which shows that the method used (real video, \sgnify, \smplifyxstar, or \pymafx) has a statistically significant effect ($p<0.001$) on recognition rate.
Pairwise comparison with Wilcoxon signed-rank tests and a Bonferroni post-hoc correction reveal that the average sign recognition rate with real video and \sgnify are both significantly higher than \smplifyxstar and \pymafx.
Importantly, sign recognition rates with real video and \sgnify are not significantly different from one another.

The central plot of \cref{fig:percres} shows the participants' perceived easiness in recognizing the sign.
These four distributions passed the normality test and were analyzed with a one-way repeated-measures ANOVA with a Bonferroni post-hoc correction for pairwise comparisons.
As expected, real videos are perceived to be significantly easier to recognize than the three reconstruction methods.
However, signs reconstructed with \sgnify and \smplifyxstar are significantly easier to recognize than those by \pymafx.

The right plot of \cref{fig:percres} shows the participants' perceived naturalness of the articulation of the signs.
All four distributions passed the normality test, so they were analyzed in the same way as perceived easiness.
All methods are statistically different one from another, with real video receiving the highest naturalness ratings, followed by \sgnify, \smplifyxstar, and \pymafx.
Nearly all participants reported in the comments that they want avatars with more expressive face motions and smoother hand and body motions.\looseness=-1

These results show that \sgnifyx outperforms the state-of-the-art, \pymafx, in all qualitative metrics.
Incorporating SL linguistic priors into our baseline \smplifyxstar yields statistically significant perceptual improvements, as judged by 20 expert signers.
Most importantly, \sgnify achieves a high sign-recognition rate that is not different from that of real-person video on the selected set of ASL signs.

A few participants suggested having clothed (rather than ``nude'') avatars and introducing more human-like features.
We thus conduct a second perceptual study to see whether these recommendations have a positive impact on the intelligibility of \sgnify's reconstructions.
Thirteen participants from the first study participated in this follow-up; the study design is the same.
The tested methods are the real video and three different \sgnify avatars: the solid purple avatar from the first study, the same avatar wearing a black long-sleeved t-shirt, and a fully textured human character adapted from Meshcapade~\cite{meshcapade} (see \supmat).
This study comprises 24 signs (including the four for participant training) from the same \asl dataset~\cite{tennant2010american} but different from those of the first study.
Our results reveal that adding the t-shirt and using a fully textured avatar do not benefit actual or perceived sign recognition or the perceived naturalness of the reconstruction.
As in the previous study, recognition of \sgnify's reconstructions were not statistically different from real video.
\section{Discussion}
Our results show that \sgnifyx performs quantitatively better than the state of the art, in particular due to the inclusion of our novel linguistic constraints.
However, we believe that a per-frame metric is not ideal for SL.
To recognize a sign, the temporal evolution is crucial, and this is not captured by \vtov.
For example, the few slightly inaccurate frames of the \smplifyxstar reconstruction of DOLL (see video on our project page) confused many signers during the perceptual study. 
Small changes over time can disrupt perception, while overall \vtov error remains small.
In the end, what matters is whether the meaning is clear to a human.
We think a perceptual study provides key insights that complement metric evaluation.
The perceptual study indicated that \sgnify significantly outperforms the state of the art and, most importantly, produces the first \threeD avatars to achieve a sign-recognition accuracy that is not statistically different from the source videos.
Our perceptual study also highlights the next challenge for SLC: the need for improvements in the face including facial expressions, tongue and eye movements, mouth morphemes, and eyebrows.
\section{Conclusions}
We present \sgnifyx, 
which estimates \threeD avatars of isolated SL signs from monocular \rgb video.
Quantitative and qualitative experiments show that
\sgnifyx outperforms the state of the art in estimating challenging SL hand poses by leveraging constraints derived from linguistics.
\sgnify represents a step towards the capture of realistic \threeD avatars from SL videos in the wild.
Future work should explore the use of our constraints in training regression methods, real-time processing, and continuous signing.

{%
\vspace{-0.7em}
\paragraph*{Acknowledgments}
{We thank Galina Henz and Tsvetelina Alexiadis for trial coordination; Matvey Safroshkin, Markus Höschle, Senya Polikovsky, Tobias Bauch, Taylor McConnell (TM), and Bernard Javot for the capture setup; %
TM for data-cleaning coordination; Leyre Sánchez Vinuela, Andres Camilo Mendoza Patino, and Yasemin Fincan for data cleaning; Nima Ghorbani and Giorgio Becherini for MoSh++; Joachim Tesch for help with Blender; Benjamin Pellkofer and Joey Burns for IT support; Yao Feng, Anastasios Yiannakidis, and Radek Dan\v{e}\v{c}ek for discussions on facial methods; Haliza Mat Husin and Mayumi Mohan for help with statistics; and the International Max Planck Research School for Intelligent Systems (IMPRS-IS) for supporting Maria-Paola Forte and Peter Kulits. 
}}

\vspace{1ex}
{
    \small
    \qheading{Disclosure}
    \href{https://files.is.tue.mpg.de/black/CoI_CVPR_2023.txt}{
         https://files.is.tue.mpg.de/black/{CoI\_CVPR\_2023.txt}}
}

{
\balance
\bibliographystyle{config_template/ieee_fullname}
\bibliography{main}
}

\clearpage

\nobalance
\appendix
\counterwithin{figure}{section}
\counterwithin{table}{section}
\counterwithin{equation}{section}
{\noindent\LARGE\textbf{Appendices}}
\setlength{\grammarindent}{2cm}
\newcommand{\indalt}[1][2]{\\\hspace*{#1em}\textbar\quad}

\section{Examples of Sign Classes}
\begin{table*}
\begin{center}
\resizebox{0.82\linewidth}{!}{
\begin{tabular}{m{4em}m{4em}cccc}
\multirow{2}{*}{\begin{tabular}[c]{@{}c@{}}~~Initial~~\\~Hand Pose\end{tabular}} & \multirow{2}{*}{\begin{tabular}[c]{@{}c@{}}~~Final~~\\~Hand Pose\end{tabular}}
& \multirow{2}{*}{~~Class~~} & \multirow{2}{*}{\begin{tabular}[c]{@{}c@{}}~~Hand-Pose~~\\~Symmetry\end{tabular}} &
\multicolumn{2}{c}{Hand-Pose Invariance} \\ 
&  & & & Dominant & Non-dominant \\
\toprule

\includegraphics[width=0.10\textwidth,clip=true,trim=280mm 100mm 300mm 100mm]{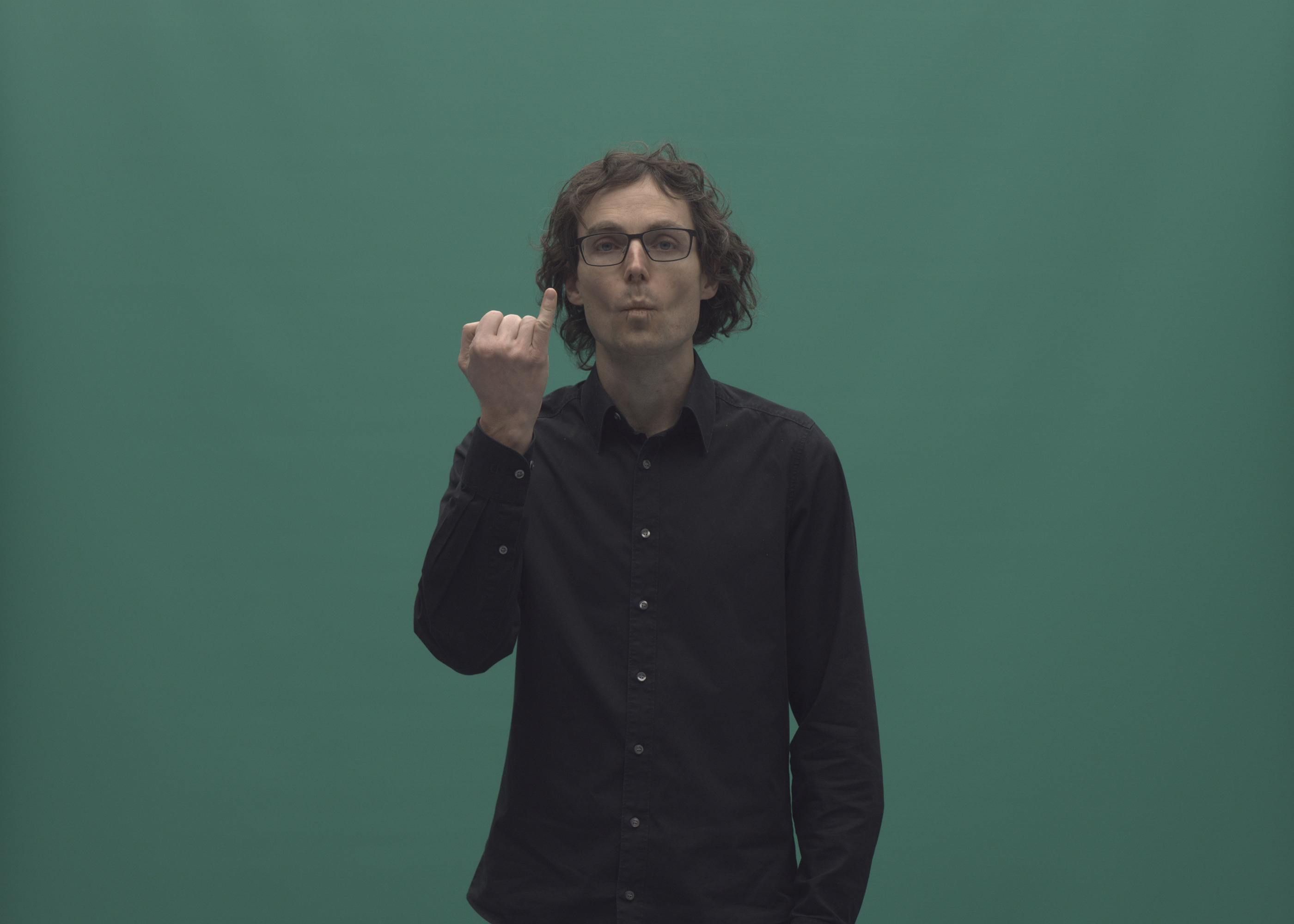} &
\includegraphics[width=0.10\textwidth,clip=true,trim=280mm 100mm 300mm 100mm]{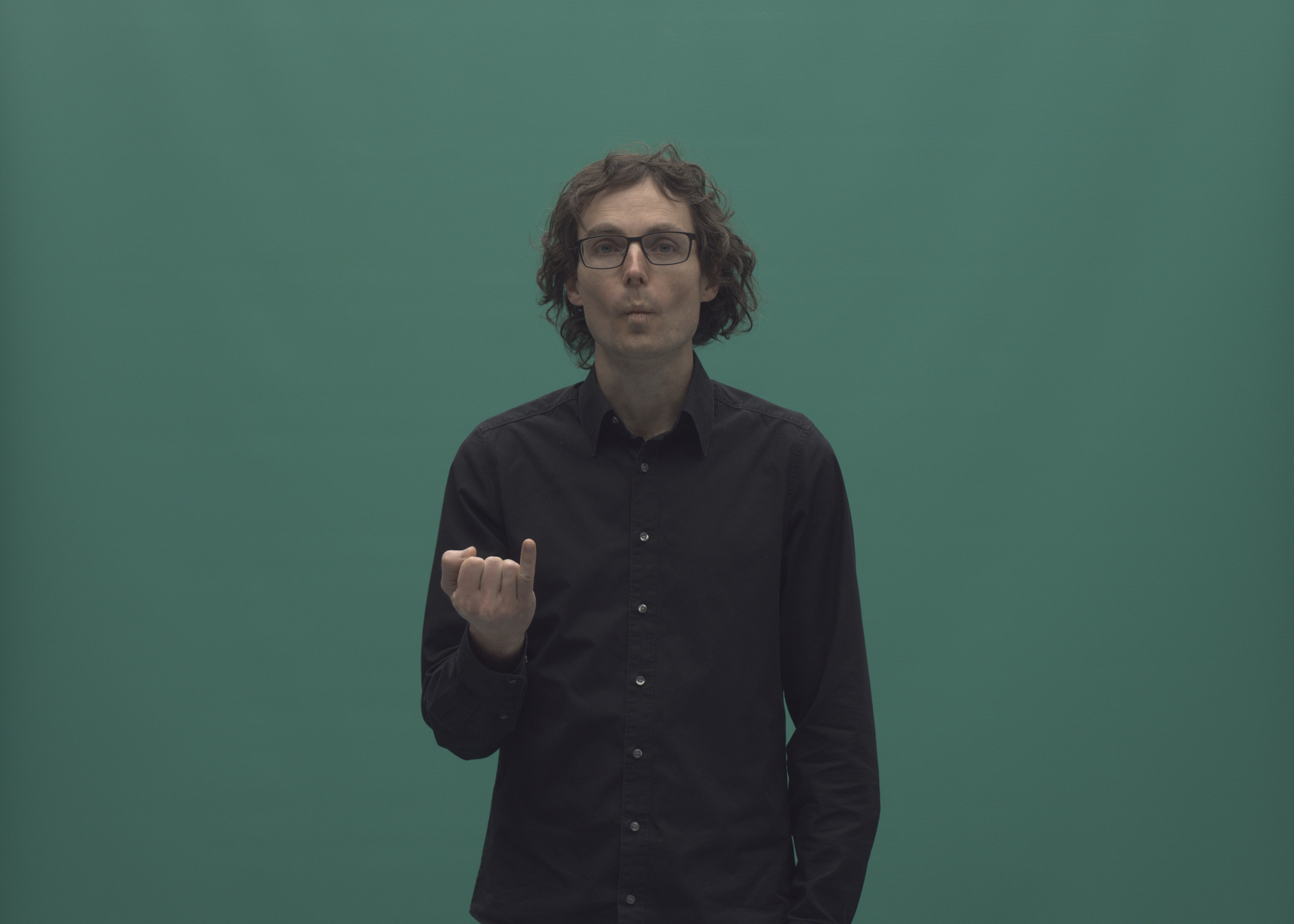}  & {0a} & \xmark & static & \xmark \\
\includegraphics[width=0.10\textwidth,clip=true,trim=240mm 100mm 280mm 100mm]{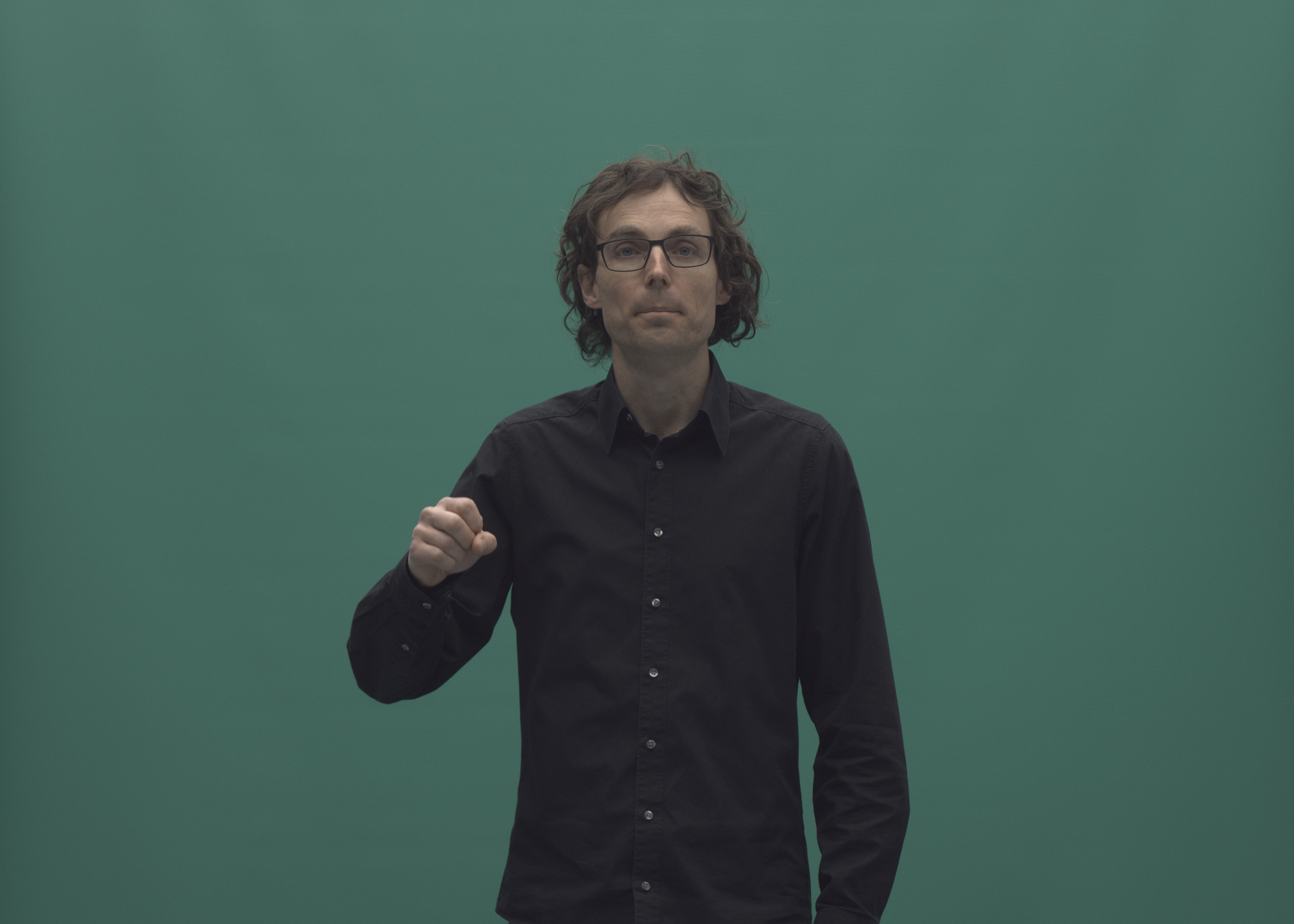} &
\includegraphics[width=0.10\textwidth,clip=true,trim=240mm 100mm 280mm 100mm]{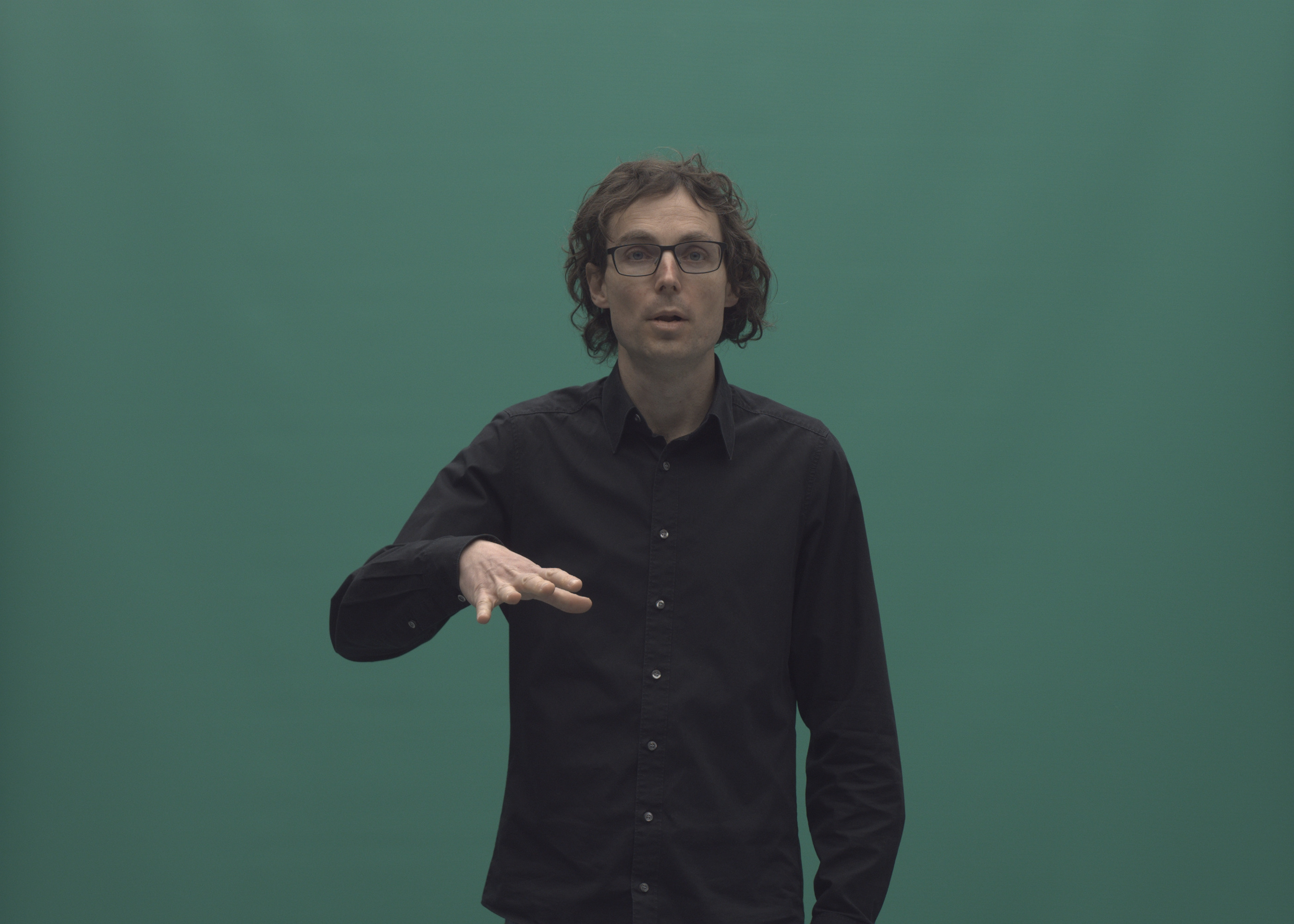}
&0b & \xmark & transitioning & \xmark \\
\includegraphics[width=0.10\textwidth, clip=true,trim=280mm 100mm 290mm 100mm]{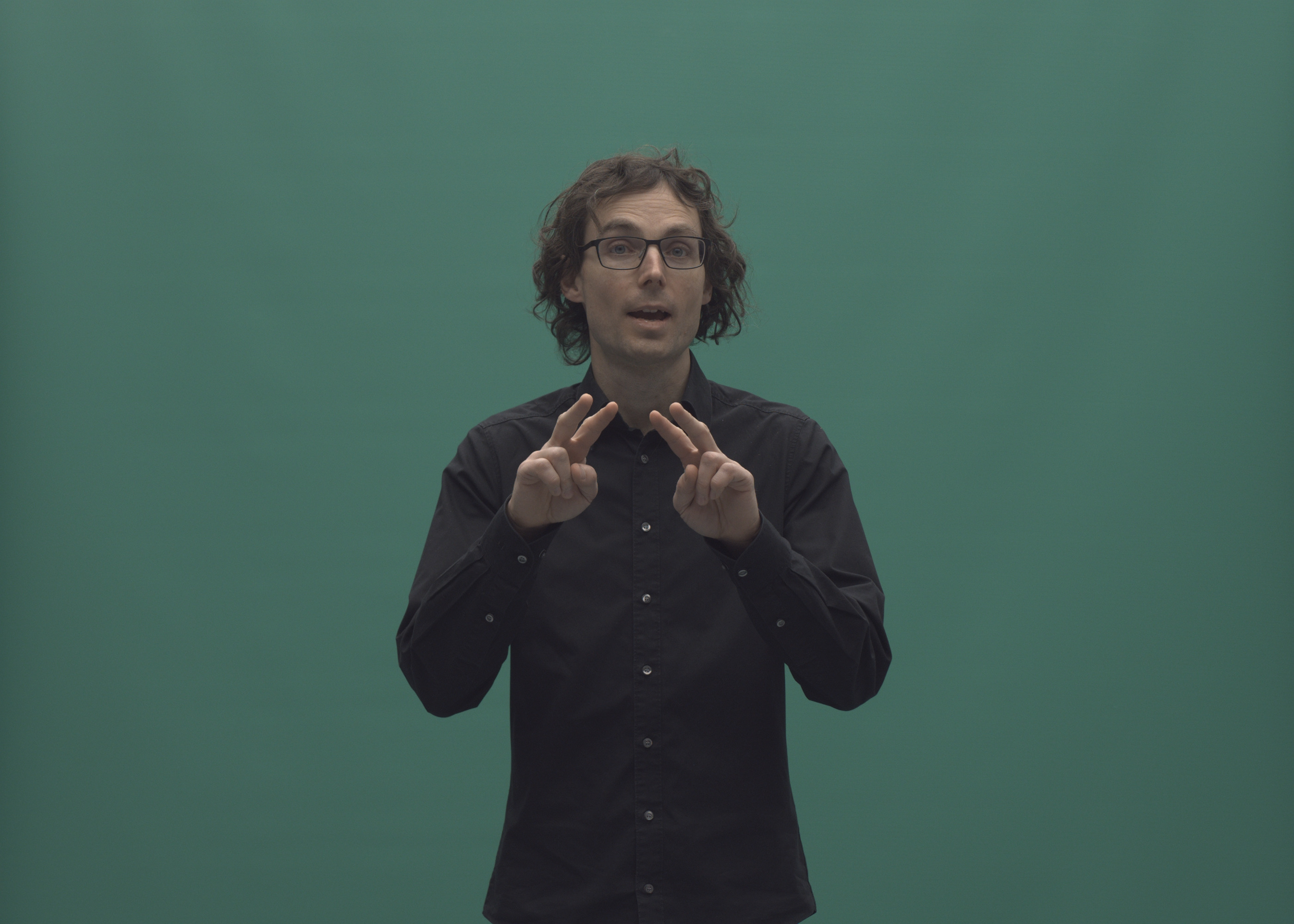} &  
\includegraphics[width=0.10\textwidth, clip=true,trim=280mm 100mm 290mm 100mm]{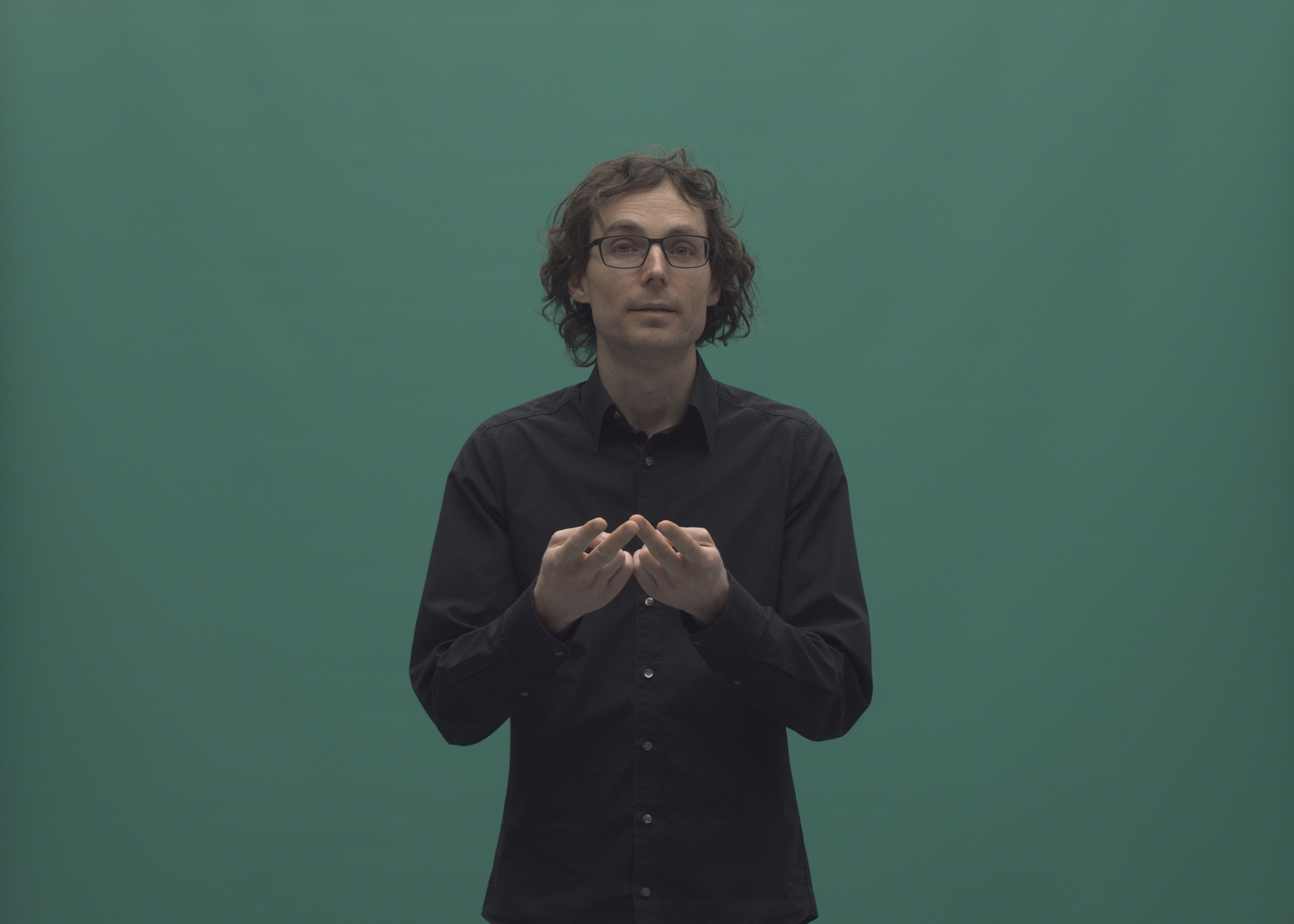}
&1a & \cmark & static & static \\
\includegraphics[width=0.10\textwidth, clip=true,trim=240mm 100mm 270mm 100mm]{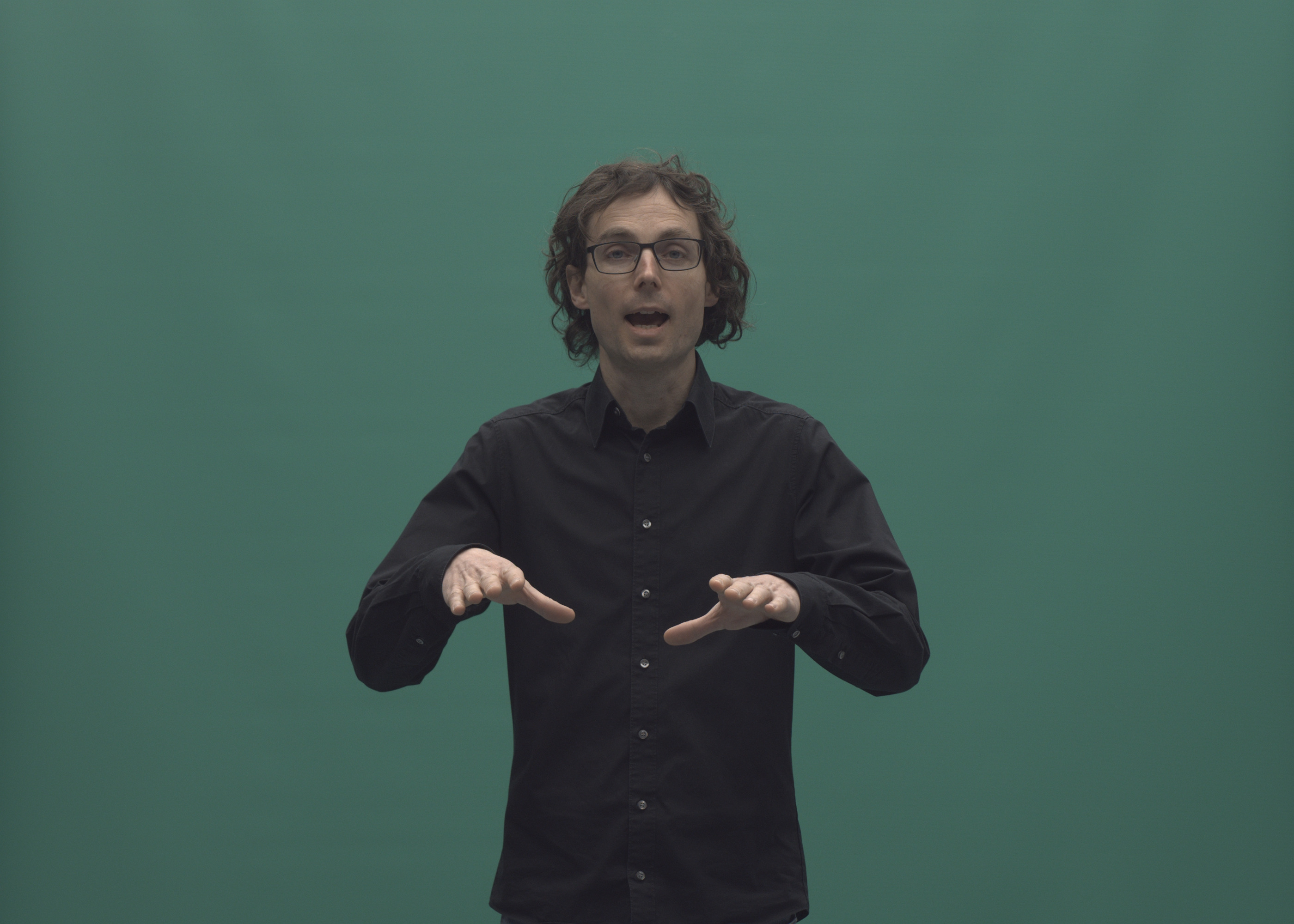} &
\includegraphics[width=0.10\textwidth, clip=true,trim=240mm 100mm 270mm 100mm]{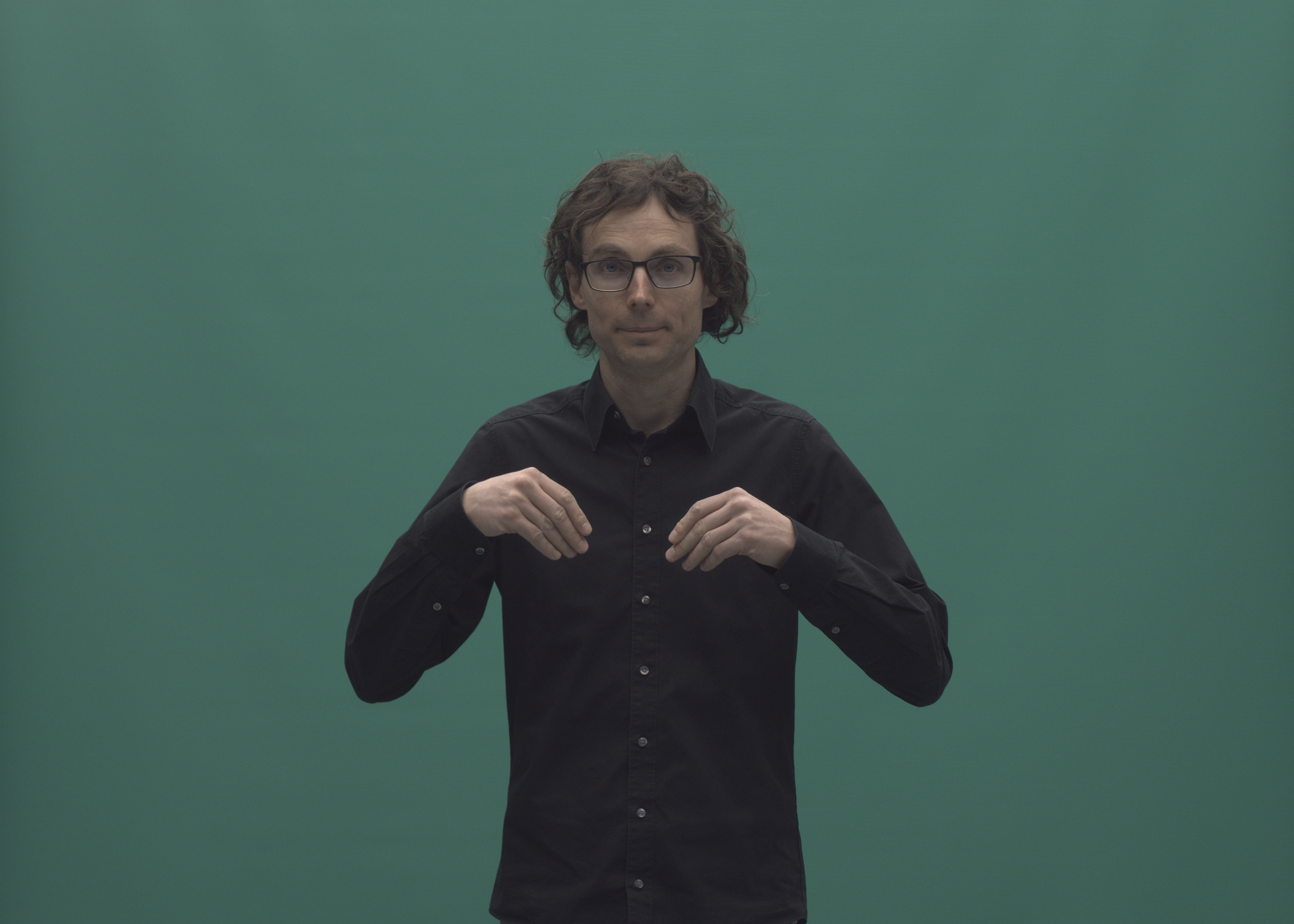}
&1b & \cmark & transitioning & transitioning \\
\includegraphics[width=0.10\textwidth,clip=true,trim=280mm 100mm 310mm 100mm]{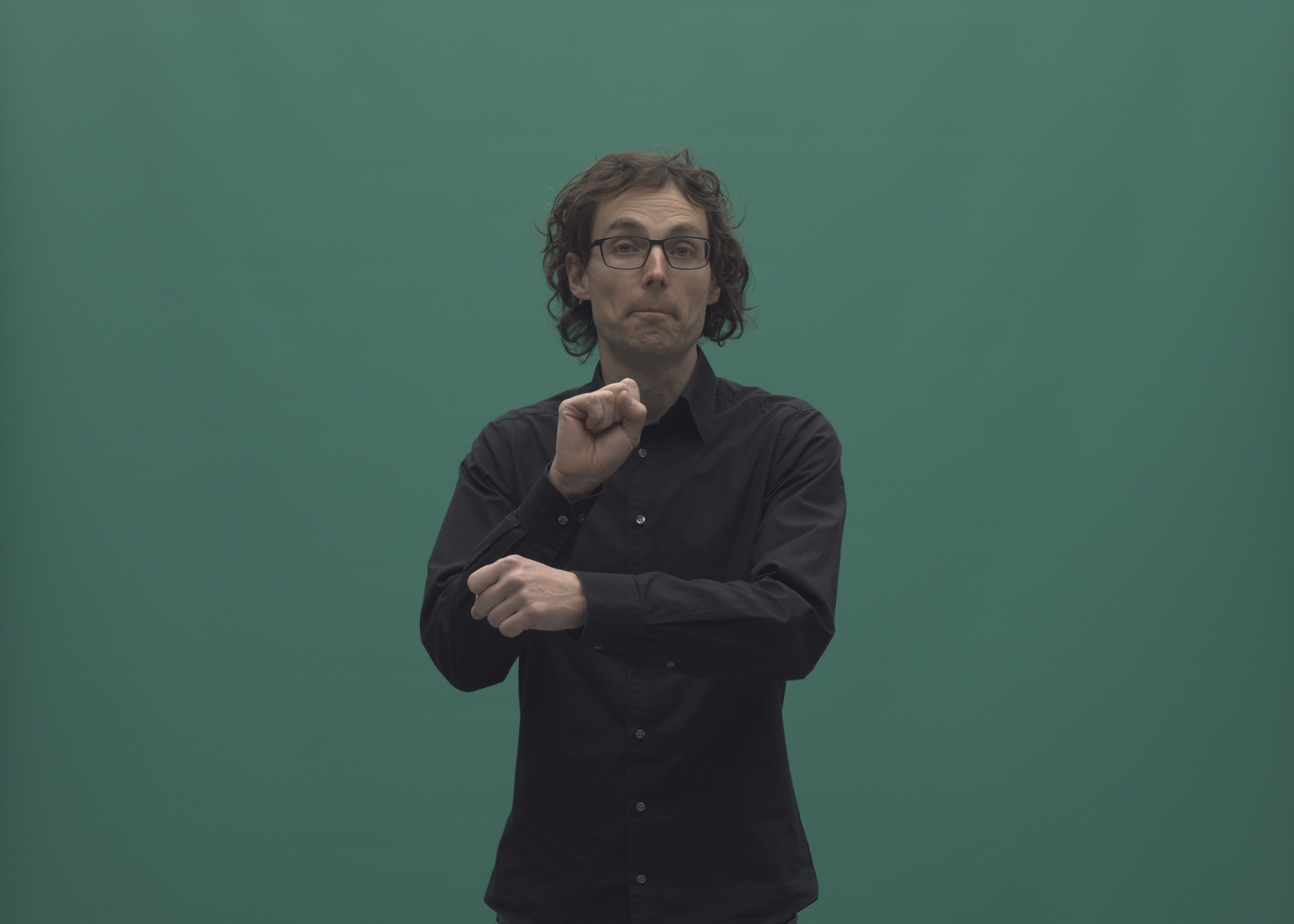} &
\includegraphics[width=0.10\textwidth,clip=true,trim=280mm 100mm 310mm 100mm]{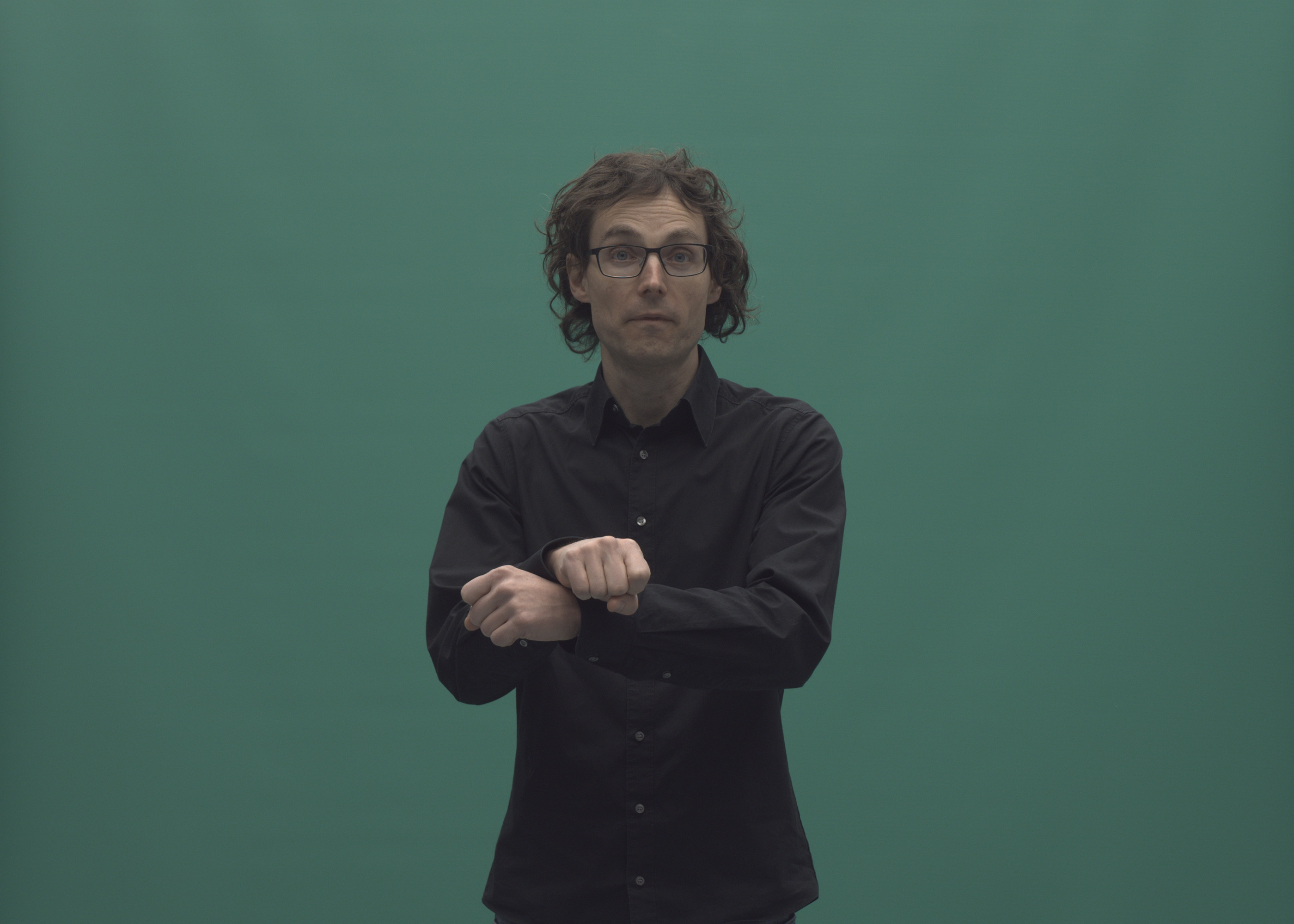}
 &2a & \cmark & static & static \\
\includegraphics[width=0.10\textwidth,clip=true,trim=280mm 100mm 310mm 100mm]{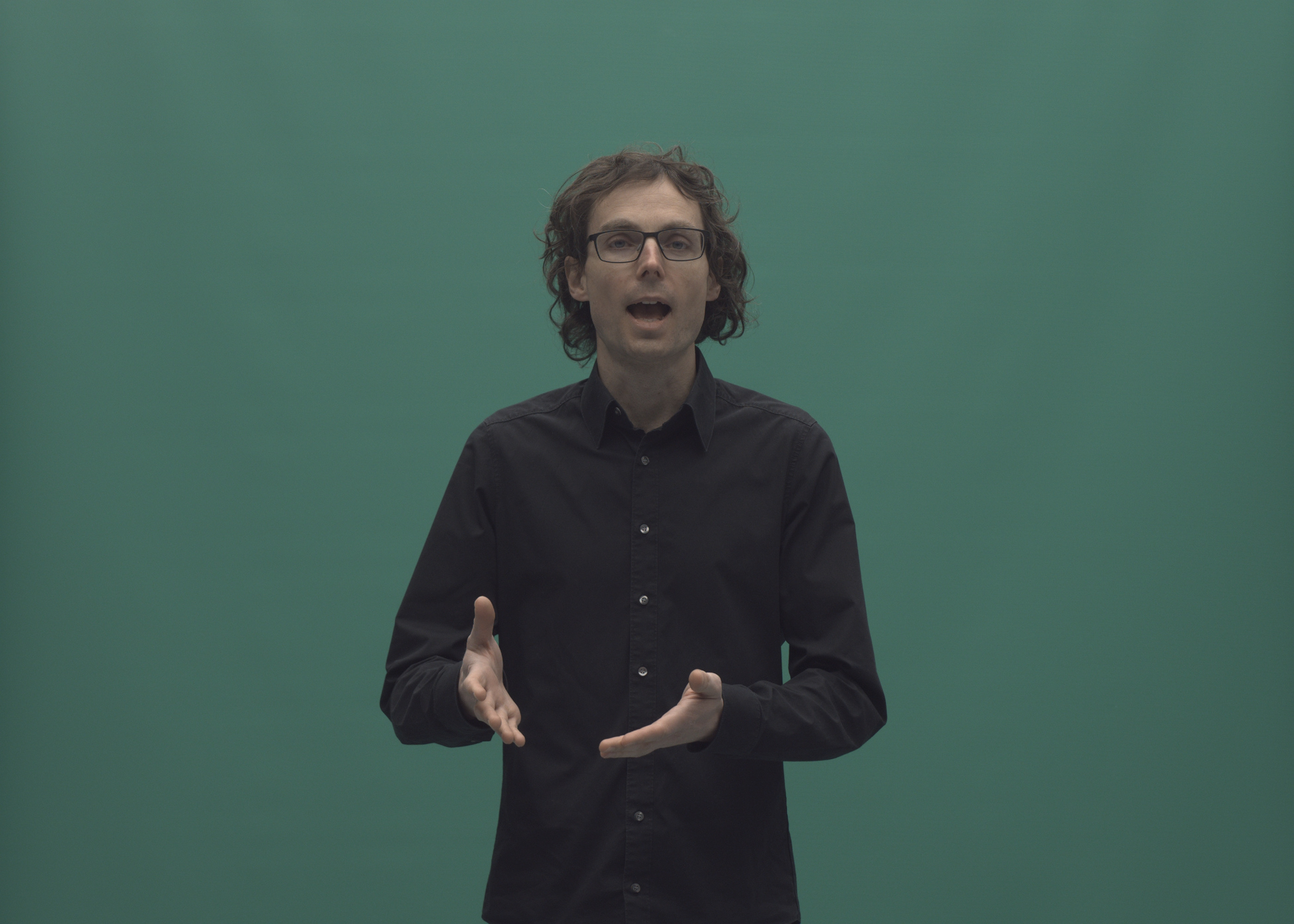} & 
\includegraphics[width=0.10\textwidth,clip=true,trim=280mm 100mm 310mm 100mm]{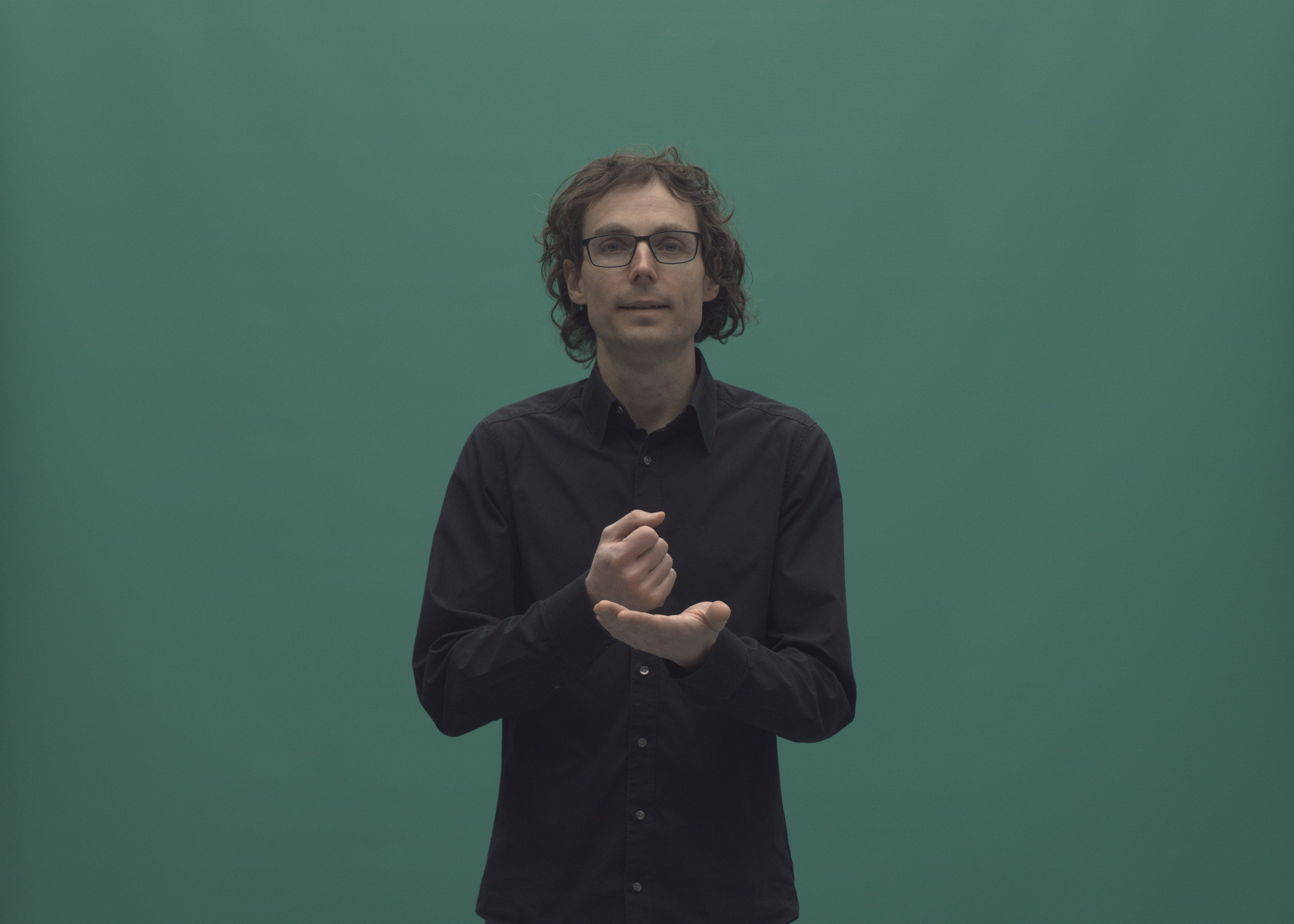}
  &2b & \xmark & transitioning & static \\
\includegraphics[width=0.10\textwidth, clip=true, trim=230mm 100mm 300mm 100mm]{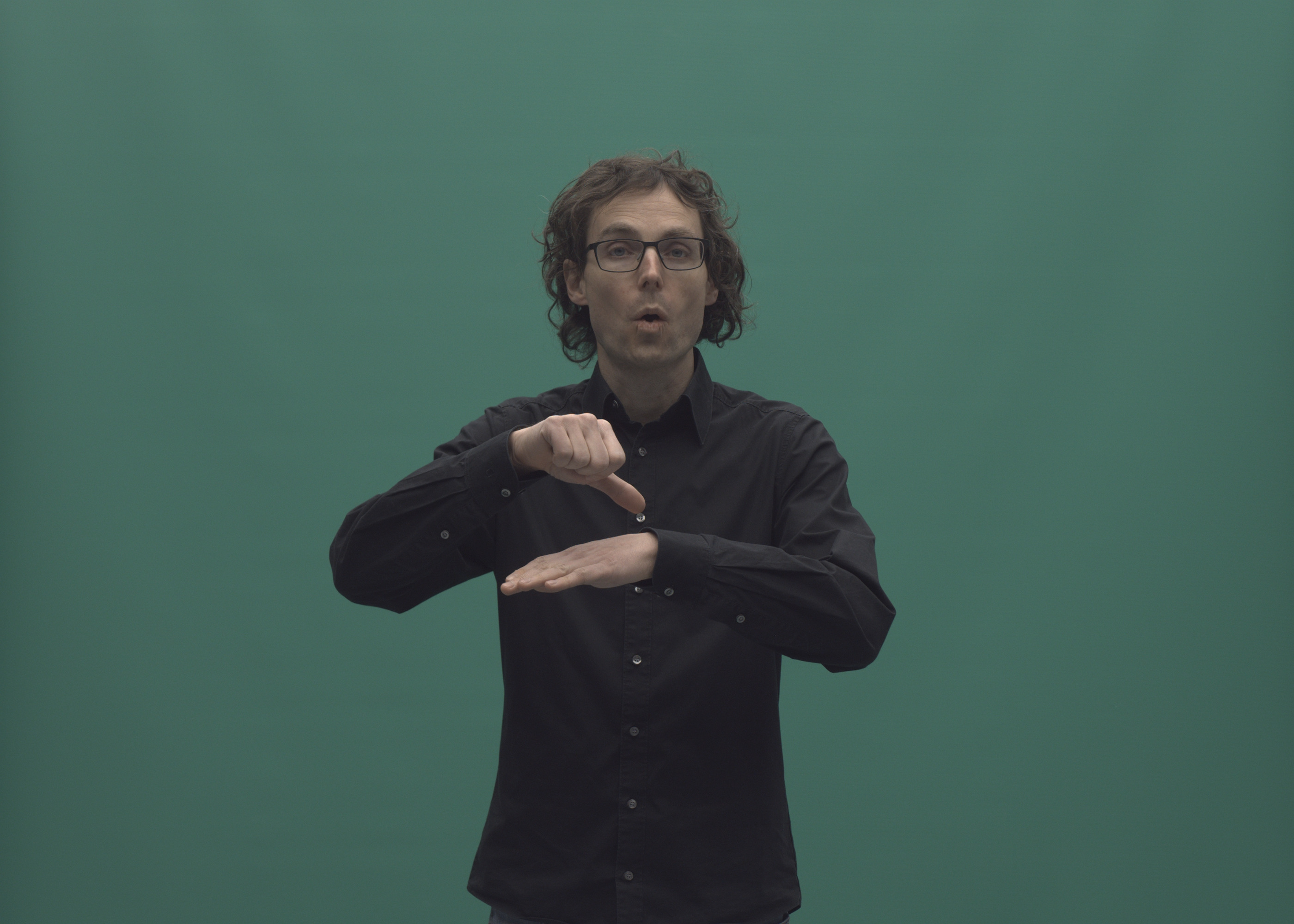} &
\includegraphics[width=0.10\textwidth, clip=true, trim=230mm 100mm 300mm 100mm]{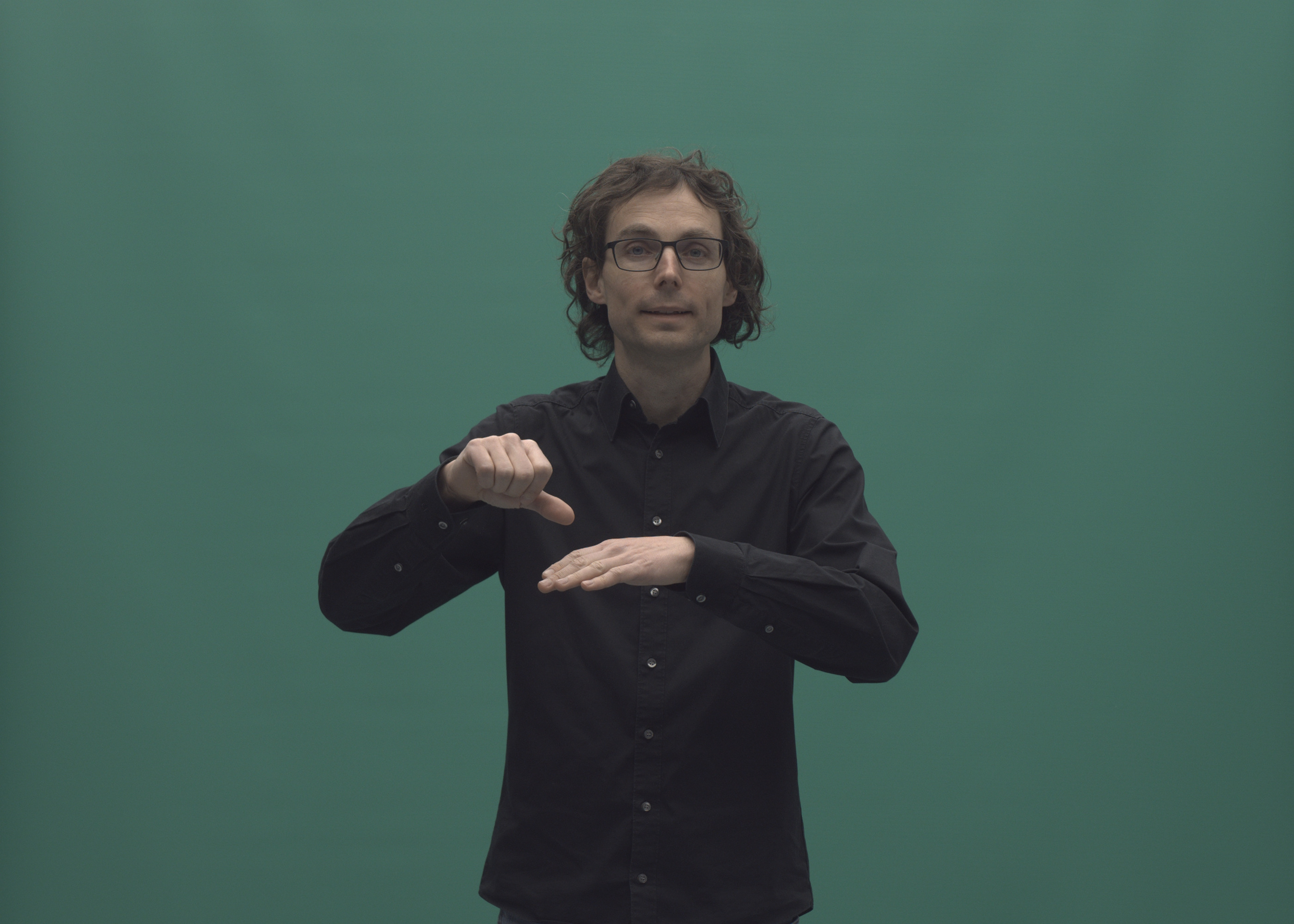}
&3a & \xmark & static & static \\
\includegraphics[width=0.10\textwidth, clip=true, trim=300mm 100mm 300mm 100mm]{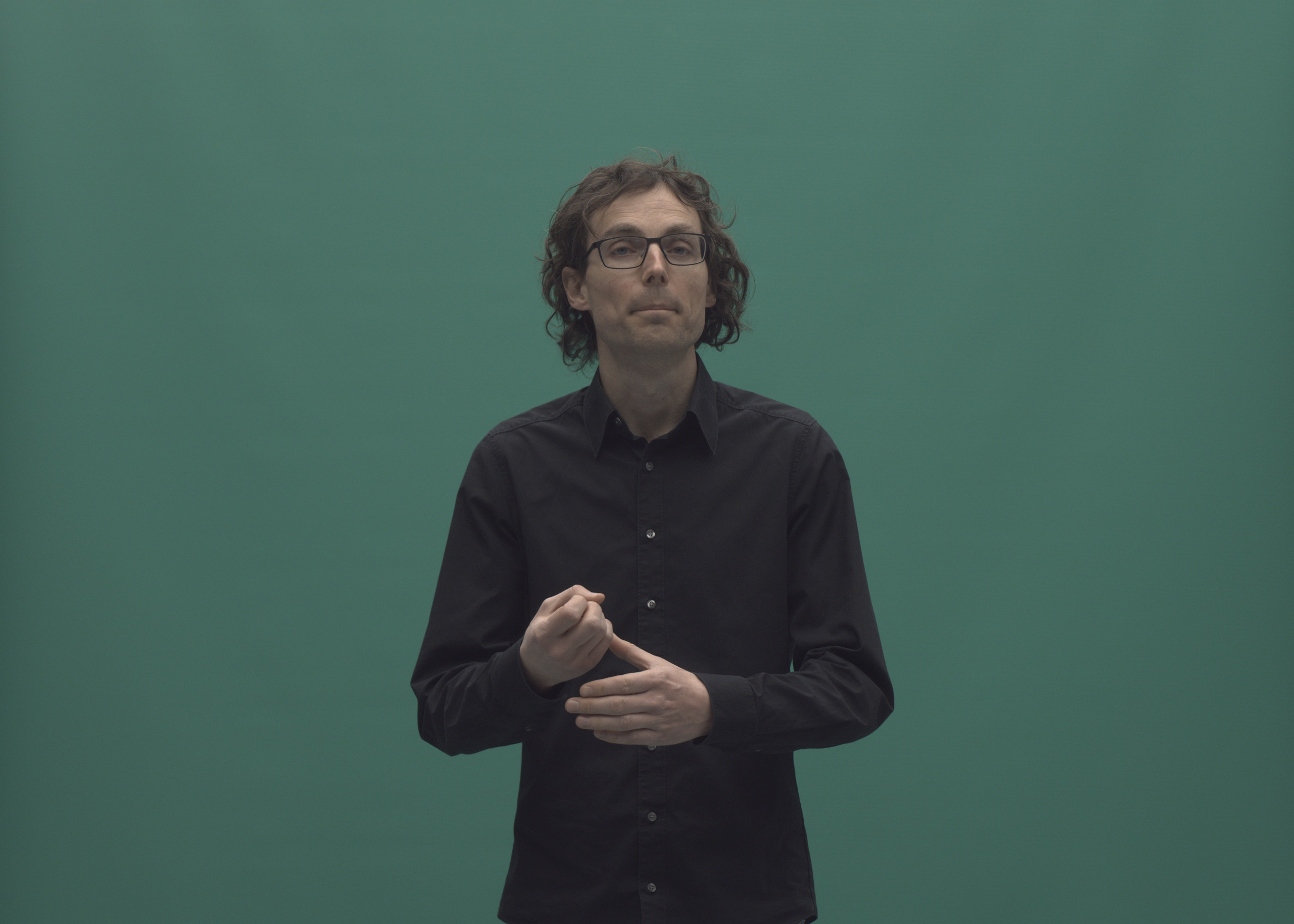} &
\includegraphics[width=0.10\textwidth, clip=true, trim=300mm 100mm 300mm 100mm]{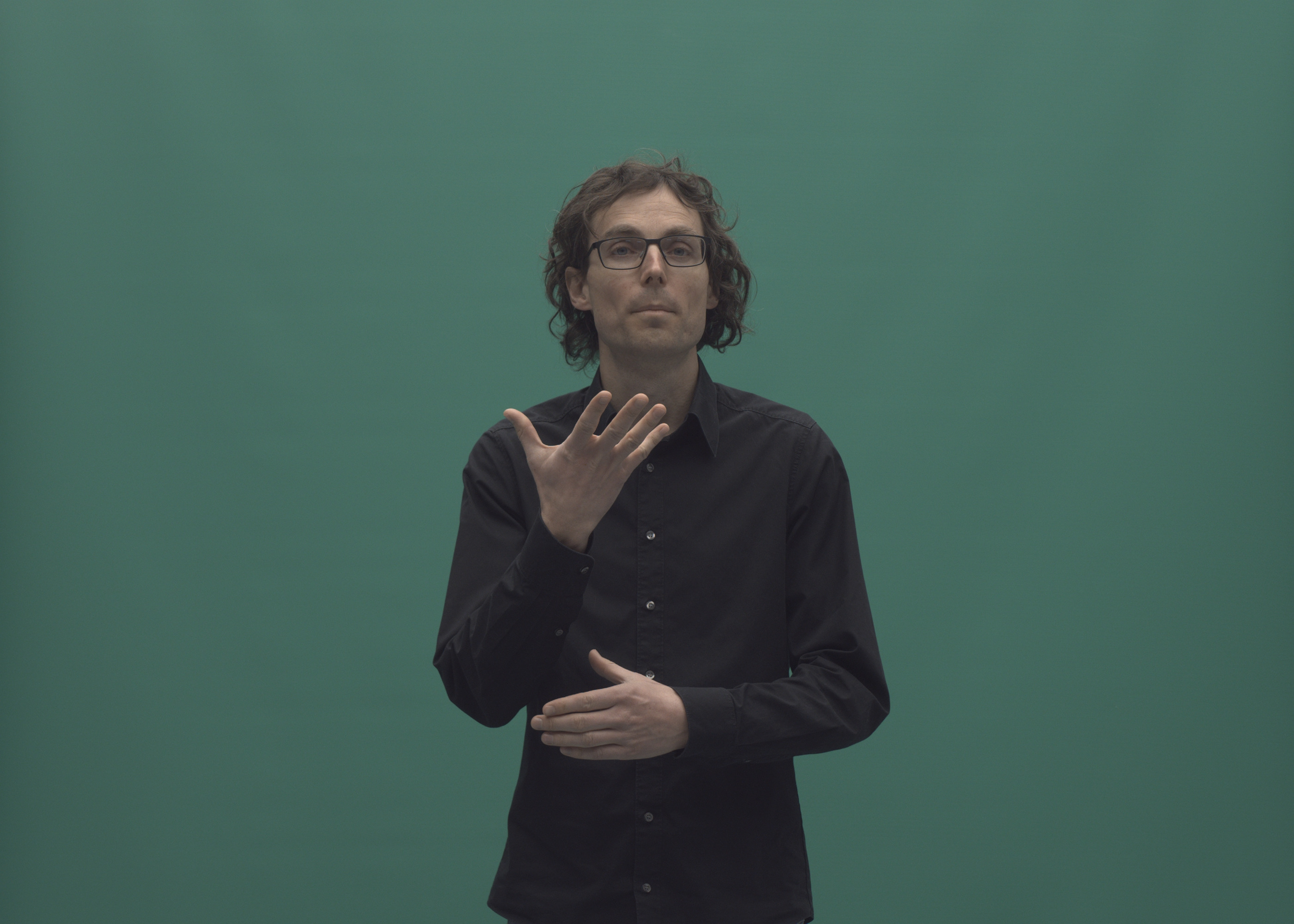}  
& 3b & \xmark & transitioning & static \\
\bottomrule
\end{tabular}
}
\end{center}
\caption{Linguistic constraints defining the eight sign classes. See supplemental video.}
\label{tab:classes_images}
\end{table*}

\Cref{tab:classes_images} provides representative images of our eight sign classes to supplement \cref{tab:classes}. 
The videos of these signs appear in the supplemental video.

\section{\sgnify Objective}

The full objective function of \sgnifyx is: 
\begin{flalign}
  E(\theta,\psi,\beta) = ~
 &  \lambda_{\theta_b} 	E_{\theta_b} +
  	\lambda_{m_h}	E_{m_h} + \nonumber\\
	&	E_J + \lambda_{\alpha}		 	E_{\alpha}
        + E_O + \nonumber\\
&  \lambda_{P} E_{P} + \lambda_A E_A + \nonumber\\
& L_s + \sum_{h\in{\{r,l\}}} L_{i}^{h} + \nonumber\\
& \lambda_t L_t + \lambda_{st} L_{st} ,
  \label{eq:objective}
\end{flalign}
where $\theta$ is the full set of optimizable pose parameters, and $\theta_b$ and ${m_h}$ are the pose vectors for the body and the two hands.
The body pose is modeled by a VAE (called Vposer) that transforms the body pose, $\theta_b$, into a latent vector $Z$.
We enforce an L2 prior in this space\ie $E_{\theta_b}(\theta_b) = \| Z \|^2$.
For the hands, \smplx uses a low-dimensional PCA pose space such that $\theta_h = \sum_{n=1}^{|m_{h}|}m_{h_{n}}\mathcal{M}$, where $\mathcal{M}$ are principal components capturing the finger pose variations and $m_{h_{n}}$ are the corresponding PCA coefficients.
Thus, $E_{m_h}(m_h)$ is an L2 prior on the coefficients $m_h$.
$E_J$ represents the joint re-projection loss, and $E_{\alpha}(\theta_b)$ is a prior penalizing extreme bending only for elbows and knees.
For more details on these terms, please refer to the original paper of \smplifyx~\cite{SMPL-X:2019}. 
$E_O$ is a bone-orientation term, which factors out the residual of the parent joint from the residual of the child joint.
For more details about this term, please refer to the original paper of RICH~\cite{RICH}.
$E_P$ and $E_A$ are used to prevent self-interpenetration. When self-contact occurs, the $E_P$ term pushes vertices that are inside the mesh to the surface, and $E_A$ aligns the surface normals of the vertices in contact. For more details, please refer to the original paper of TUCH~\cite{muller2021self}.

We added $L_s$ and $L_{i}^{h}$ to enforce our linguistic constraints: $L_s$ represents the symmetry constraints, and $L_{i}^{h}$ the hand-pose invariance of the right ($r$) and left ($l$) hands, as described in \cref{sec:lingconstr}.
We also added a temporal loss $L_t$ on the body- and hand-pose vectors and a standing loss $L_{st}$ to penalize deviations from a standing pose when none of the feet keypoints are detected; specifically, this penalization is applied to the joints below the pelvis and to the spine.

Finally, each $\lambda$ denotes the influence weight of each loss term.
For more details on the exact $\lambda$ values and insights on the full \sgnify objective,
please see the code, which can be reached from the project URL.

We optimize our objective function using the trust-region Newton conjugate gradient method~\cite{NoceWrig06}.
Note that we do not optimize for the shape $\beta$ and the facial expressions $\psi$.

\begin{figure}[b]
\centering
   \includegraphics[width=\linewidth,clip=true,trim=20mm 46mm 45mm 15mm]{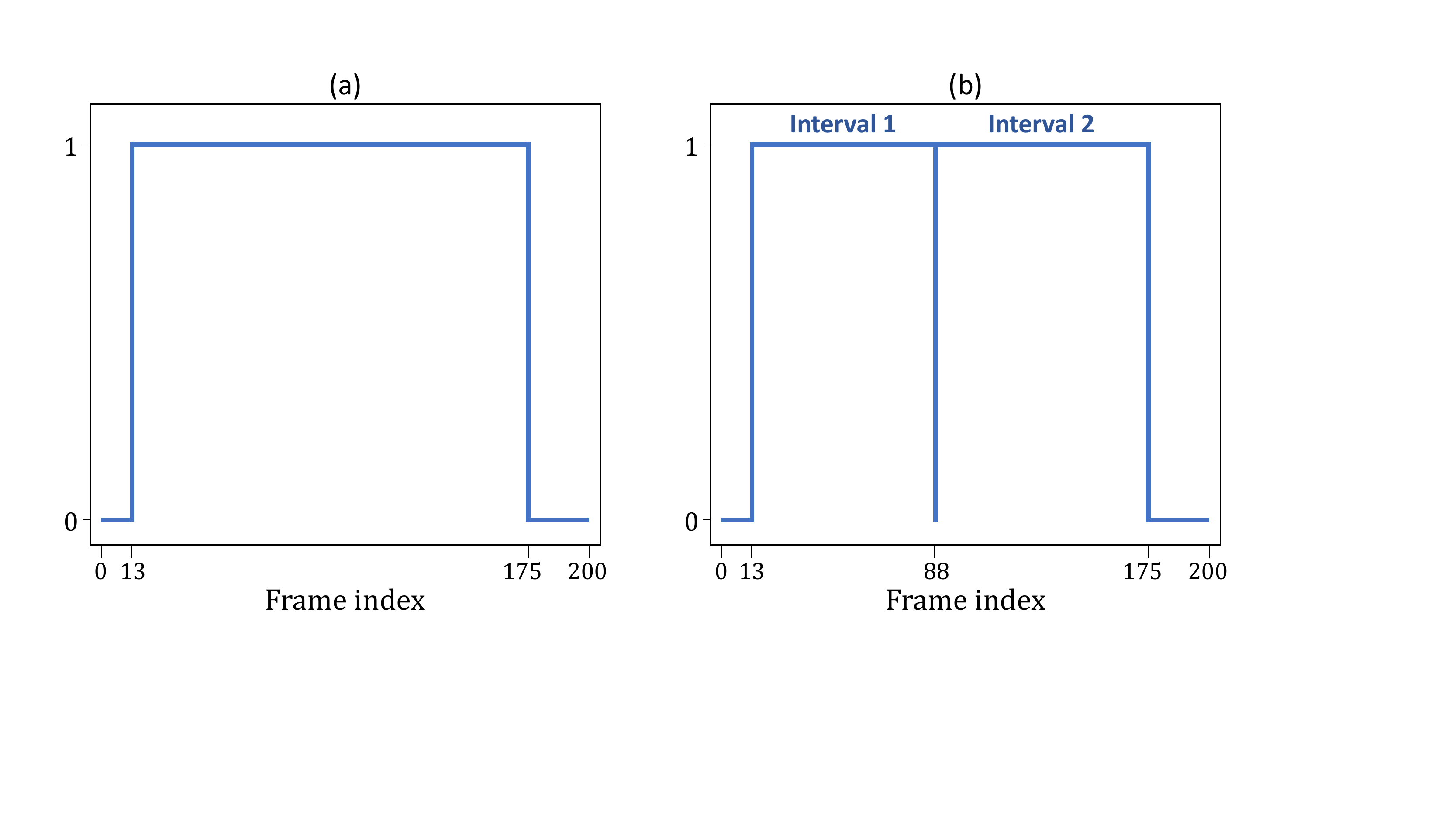}
\caption{We consider an example sequence of 200 frames. 
(a) Static hand: Frames 
whose value on the y-axis is 1 are candidates for identifying $\thetarefhand$.
(b) Transitioning hand and input features for the sign-group classifier: The first interval shows candidates for $\thetarefhandone$, and the second one for $\thetarefhandtwo$.}
\label{fig:interval}
	
\end{figure}
\begin{figure*}[hbt!]
\centering
\begin{framed}
\begin{grammar}
  <hns> ::= [SYMMETRY] <block>
  
  <block> ::= [<handshape_block> | <non_handshape_block>]*
  
  <handshape\_block> ::= HANDSHAPE [HANDSHAPE\_MODIFIER | HANDSHAPE\_FINGER\_LOCATION]*
  
  <non\_handshape\_block> ::= <par> | <seq> | <fusion> | EXTENDED\_FINGER\_LOCATION | PALM\_ORIENTATION | MOVEMENT | MOVEMENT\_MODIFIER | LOCATION | LOCATION\_MODIFIER | OTHER\_SYMBOL\_\_NO\_GROUP

  <par> : HAMPARBEGIN <block> [HAMPLUS <block>] HAMPAREND

  <seq> : HAMSEQBEGIN <block> HAMSEQEND

  <fusion> : HAMFUSIONBEGIN <block> HAMFUSIONEND
\end{grammar}
\end{framed}
\caption{
Constructed HamNoSys EBNF grammar.
}
\label{fig:ebnf}
\end{figure*}

\section{Intervals for Selecting the Candidate Frames for the Reference Hand Poses ($\thetarefhand$, $\thetarefhandone$, and $\thetarefhandtwo$)}
When articulating an isolated sign, signers start and end in a rest pose.
\sgnifyx identifies the beginning and end of the sequence based on when the hands begin to move.
After automatic trimming, the initial and final frames of the sequence show the transition from the rest pose to the pose(s) characteristic of the sign.
We observe that the transition from the rest pose to the core part of the sign usually happens around $t=0.5 * T/8$, and the transition from the sign to the rest pose typically occurs around $t=7 * T/8$, where $T$ is the number of frames in the motion sequence.
As a result, we assume the core part of a sign to happen between $0.5 * T/8<t<7 * T/8$.
\Cref{fig:interval}\textcolor{red}{a} shows the frames during which the transitions from/to the rest pose happen (indicated with 0) and the frames during which the sign is articulated (indicated with 1) for a sample trimmed recording containing 200 frames.
To identify the two key poses representing the initial and final hand poses ($\thetarefrightone$ and $\thetarefrighttwo$), we consider two different intervals; we expect to see the first hand pose at the beginning of the sequence (first interval shown in \cref{fig:interval}\textcolor{red}{b}) and the second hand pose at the end (second interval shown in \cref{fig:interval}\textcolor{red}{b}).

\begin{figure*}[t]
\vspace{2em}
\centering
\includegraphics[width=\linewidth,trim=0.5cm 1cm 0.8cm 1cm,clip]{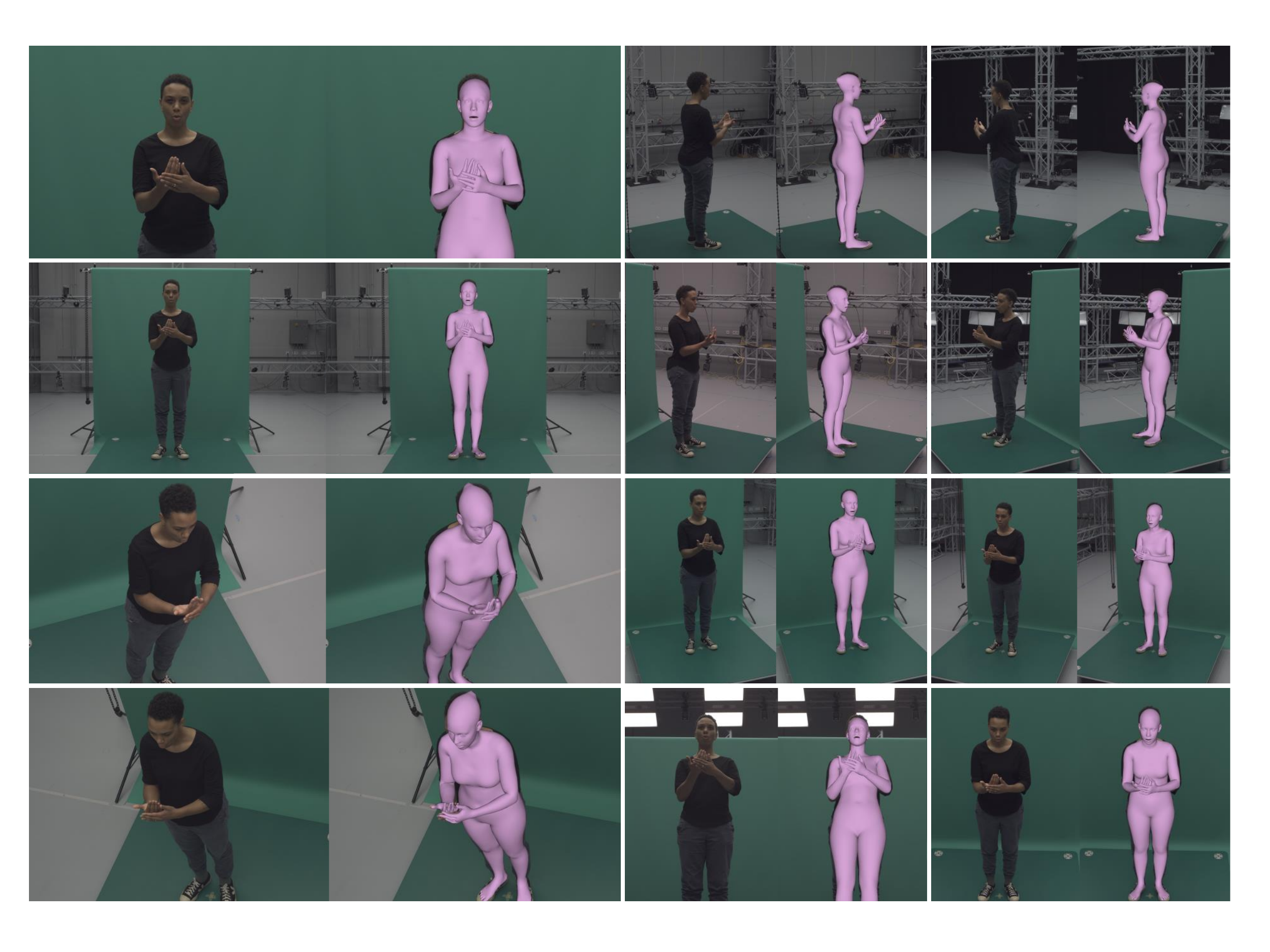}
\caption{
The multi-view setup comprises 12 synchronized \rgb cameras. A close-up frontal camera is zoomed in to focus on the hands and face.
Another frontal camera captures the entire front of the body.
Two top-lateral cameras acquire images with a top-down view.
Four lateral cameras are placed at hip level and capture the whole body; two are slightly behind the signer, and the other two are slightly in front.
Two frontal-lateral cameras also have a full-body view, looking slightly down.
Finally, two other frontal cameras, one with a bottom-up view and one with a top-down view, are focused on the hands.
The participant stands on a $1.5$~m $\times$ $1.5$~m platform of adjustable height located in front of a green screen.}
\label{fig:mv}
\end{figure*}
\begin{figure*}[t]
\vspace{4em}
\centering
    \includegraphics[width=0.98\linewidth]{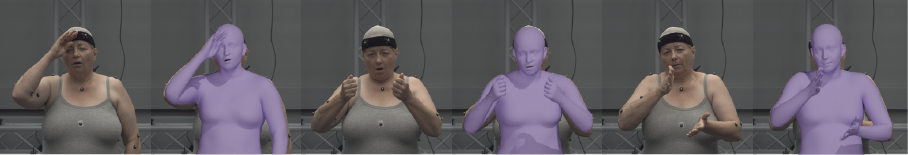}
    \vspace{-0.35em}
    \makebox[0.33\linewidth]{\small[VATER]}
    \makebox[0.33\linewidth]{\small[AUTO]}
    \makebox[0.33\linewidth]{\small[AUSGEBEN]}
\caption{Sample frames and reconstructions from segments of the German sentence: \textit{Der \textbf{Vater} muss f{\"u}r die Reparatur seines \textbf{Autos} viel Geld \textbf{ausgeben}.}}
\label{fig:sentence}
\end{figure*}

\section{HamNoSys Parsing}
We construct an Extended Backus--Naur form (EBNF) grammar (see \cref{fig:ebnf}) to parse HamNoSys~\cite{hanke2004hamnosys} annotations to a form where we can extract labels to train our sign group classifier.
HamNoSys is a universal sign-language phonetic transcription system that can be used to represent all hand poses and movements that constitute a sign; \emph{i.e.,} someone reading a HamNoSys annotation would be able to fully reproduce the sign it represents.
We parse these transcriptions on the annotated Corpus-Based Dictionary of Polish Sign Language (CDPSL)~\cite{slownikpjm}, and we assign our classes to the clips as follows:

\qheading{Class 0a} There is one \textit{handshape\_block} nonterminal and no SYMMETRY terminal is present.

\qheading{Class 0b} There are two \textit{handshape\_block} nonterminals, the two \textit{handshape\_block} nonterminals are not equal, a HAMREPLACE terminal is present, and no SYMMETRY or REPEAT terminals are present.

\qheading{Class 1a} There is one \textit{handshape\_block} nonterminal and a SYMMETRY terminal is present.

\qheading{Class 1b} There are two \textit{handshape\_block} nonterminals, they are not equal, a HAMREPLACE terminal is present, and no SYMMETRY or REPEAT terminals are present.

\qheading{Class 2a} There are two \textit{handshape\_block} nonterminals, they are equal, they fall within a \textit{par} nonterminal, and no SYMMETRY terminal is present.

\qheading{Class 2b} There are three \textit{handshape\_block} nonterminals, the first two are equal, a HAMREPLACE terminal is present, and no SYMMETRY or REPEAT terminals are present.

\qheading{Class 3a} There are two \textit{handshape\_block} nonterminals, they are not equal, they fall within a \textit{par} nonterminal, and no SYMMETRY terminal is present.

\qheading{Class 3b} There are three \textit{handshape\_block} nonterminals, the first is not equal to the second, a HAMREPLACE terminal is present, and no SYMMETRY or REPEAT terminals are present.

Note that the SYMMETRY parameter from HamNoSys refers to Battison's symmetry condition~\cite{battison1978lexical}, which also includes the signer's arm movement and not only the hand pose; in contrast, our symmetry constraint applies only to hand pose.

\section{\sgnifyx Extensions}
\subsection{Multi-view}\label{sec:mv}

If multi-view video is available, \sgnifyx is easily extended to this case.
We used $12$ synchronized \rgb cameras (see \cref{fig:mv}) at $90$ \fps to capture the same participant used in the quantitative evaluation plus two additional signers, a native signer and an interpreter with 17 years of experience.
Each participant articulated all signs in our German Sign Language (DGS) corpus (see \cref{sec:dataset}).
A close-up frontal camera is zoomed in to focus on the hands and face  of the signer and has a view similar to existing sign-language videos.
Another frontal camera captures the whole front body of the participant.
Two top-lateral cameras acquire images with a top-down view.
Four lateral cameras are placed at hip level and capture the whole body; two are slightly behind the signer, and the other two are slightly in front.
Two frontal-lateral cameras also have a full-body view, looking slightly down.
Finally, two other frontal cameras, one with a bottom-up view and one with a top-down view, are focused on the hands.
The participant stands on a $1.5$~m $\times$ $1.5$~m platform of adjustable height located in front of a green screen. 
Then, multi-view \sgnifyx is used to fit \smplx.
We follow Huang \etal \cite{RICH} to combine the keypoint predictions of different cameras.
A person-specific $\beta$ is obtained with a \threeD scanner.
Sample multi-view results are shown in the supplemental video.

\begin{figure*}[t]
\vspace{+1em}
\begin{center}
\resizebox{0.98\linewidth}{!}{
~\includegraphics[width=0.24\linewidth,trim=11mm 0mm 0mm 0mm]{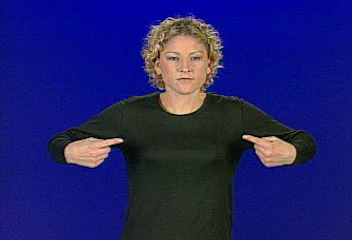}
\includegraphics[width=0.24\linewidth]{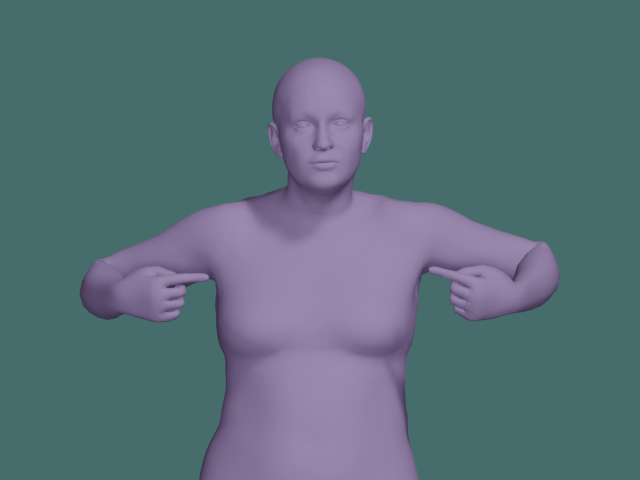}
\includegraphics[width=0.24\linewidth]{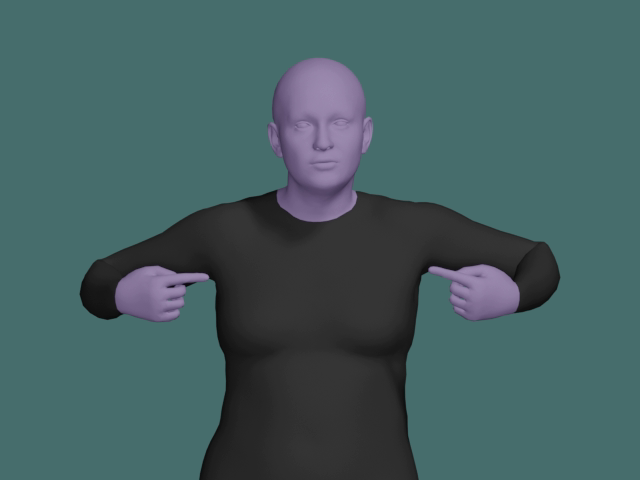}
\includegraphics[width=0.24\linewidth]{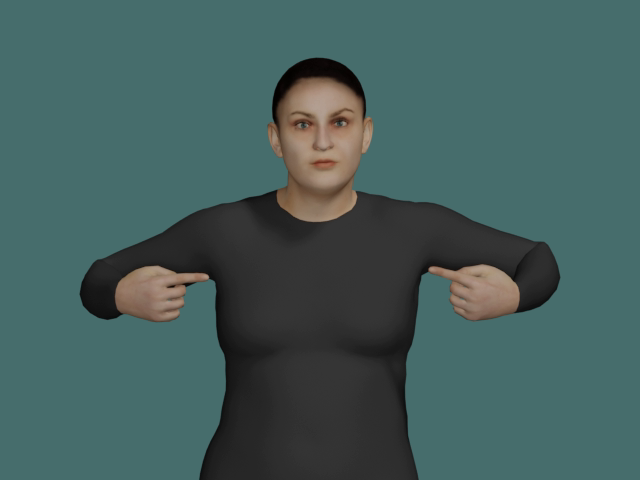}
}
\end{center}
\vspace{-1em}
\caption{Sample frames from the four methods presented in the second perceptual study: real video, the solid purple avatar from the first study, the same avatar wearing a black long-sleeved t-shirt, and a fully textured human character.
}
\label{fig:second_study}
	
\end{figure*}
\begin{table}[t]
\mytablesetup
\centering
  \resizebox{0.99\linewidth}{!}{
\begin{tabular}{ccccc}
    \toprule
        \textbf{Method} & \textbf{\upperbodylabel} & \textbf{Left Hand} & \textbf{Right Hand} \\
        \midrule 
            \frankmocap~\cite{rong2021frankmocap} & 74.93 & 23.70 & 19.57  \\
            \pixie~\cite{PIXIE:3DV:2021}  & 59.09 & 24.79 & 20.19 \\
            \pymafx~\cite{zhang2022pymaf}  & 68.30 & 22.51 & 18.49  \\ 
            \smplifyxstar & 55.71 & 21.14 & 18.60  \\
            \sgnifyx & \textbf{54.72} & \textbf{20.28} & \textbf{17.44}  \\
        \bottomrule
    \end{tabular}
    
}
\vspace{\vspaceTABaboveCaption}
\caption{Mean TR-V2V error (mm) on fluid sentences.}
    \label{tab:sent}
\end{table}
\begin{figure}[t]
\centering
\includegraphics[width=0.48\linewidth,trim= 0 90 0 100,clip]{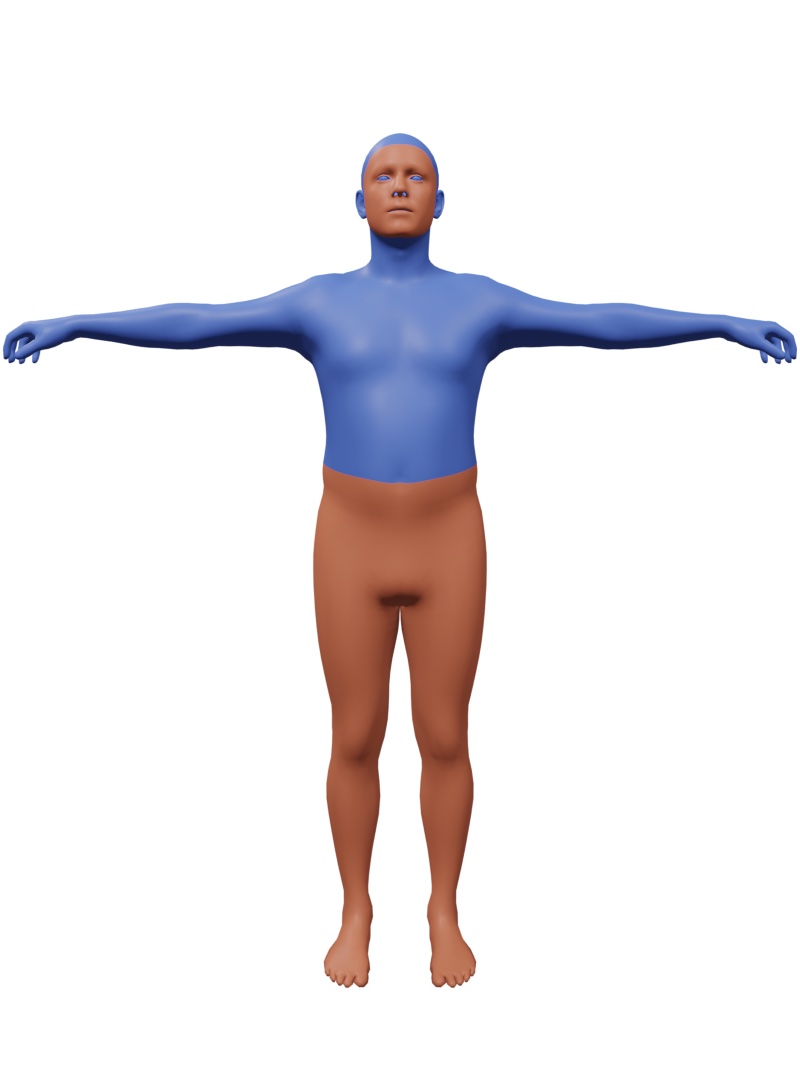}~~
\includegraphics[width=0.48\linewidth,trim= 0 90 0 100,clip]{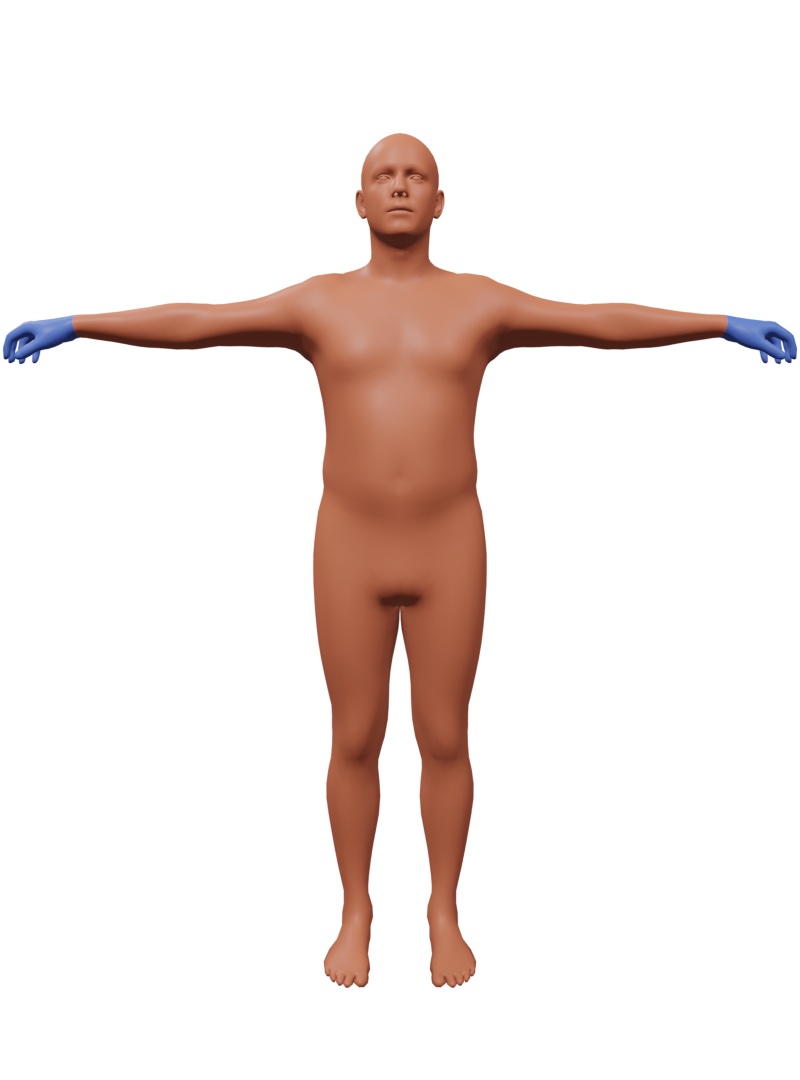}

\caption{%
Blue vertices are used to calculate vertex error metrics, while red vertices are ignored.
The left image shows the vertices used for the column of quantitative results labeled ``\upperbodylabel''\ie upper-body vertices.
The right image shows the vertex subsets for the left and right hands. Best viewed in color.
}
\label{fig:quant_eval_verts}
	
\end{figure}

\subsection{\cslclong (\cslc)}

\sgnifyx can also be used for \cslc.
Besides isolated signs, our corpus contains ten sentences articulated by the three interpreters during the four sessions; one session with a 54-camera \vicon \mocap system at $120$ \fps synchronized with a frontal $4112\times3008$ \rgb camera at $60$ \fps\/, framing an upper-body view as typically found in SL video (see \cref{sec:dataset}) and three sessions with the multi-view setup (see \cref{sec:mv}).
Depending on the interpreter, different DGS versions of the same German sentences were proposed.

We conduct an exploratory quantitative study with twelve sentences (ten main sentences and two variations) collected as in \cref{sec:dataset} and analyzed as in \cref{sec:experiment}. 
\Cref{tab:sent} shows the mean \TRvtov error across the twelve sentences for four methods and three body regions.
This experiment compares \sgnifyx with \frankmocap~\cite{rong2021frankmocap}, \pixie~\cite{PIXIE:3DV:2021}, \pymafx~\cite{zhang2022pymaf}, and our baseline \smplifyxstar. 
\sgnifyx achieves the lowest error for the upper body and both hands, beating the state-of-the-art methods.
It is interesting to notice that while \frankmocap has a hand-pose error lower than \pymafx in our previous quantitative experiment (see \cref{sec:experiment}), this is not true in this second experiment.
This inconsistency further emphasizes the limitations of a per-frame metric for sign language.
In the future, a perceptual study should be conducted to evaluate the recognition of the reconstructed sentences with proficient signers.
Such an experiment will give more insights about the next crucial steps for \cslc.
\Cref{fig:sentence} shows sample frames and \sgnifyx's reconstructions from a sentence of this exploratory study. 

\section{Vertices for Quantitative Analysis}
\Cref{fig:quant_eval_verts} illustrates the subsets of vertices selected for the quantitative evaluation.

\section{Second Perceptual Study}
\Cref{fig:second_study} shows a sample frame represented with each of the four methods used in the second perceptual study: real video, the solid purple avatar from the first study, the same avatar wearing a black long-sleeved t-shirt, and a fully textured human character adapted from Meshcapade~\cite{meshcapade}.

\section{Additional Examples}
\Cref{fig:examples1} shows additional examples from the Real SASL~\cite{reaslsasl} and CDPSL~\cite{slownikpjm} datasets.
\Cref{fig:examples2} shows additional examples from The American Sign Language Handshape Dictionary~\cite{tennant2010american} and our collected DGS dataset (see \cref{sec:dataset}).

\twocolumn[{
    \renewcommand\twocolumn[1][]{#1}
    \maketitle
    \centering
    \vspace{-0.2em}
    \begin{minipage}{1.00\textwidth}
    \centering                                 
    \includegraphics[width=\linewidth,clip=true,trim=000mm 000mm 000mm 000mm]{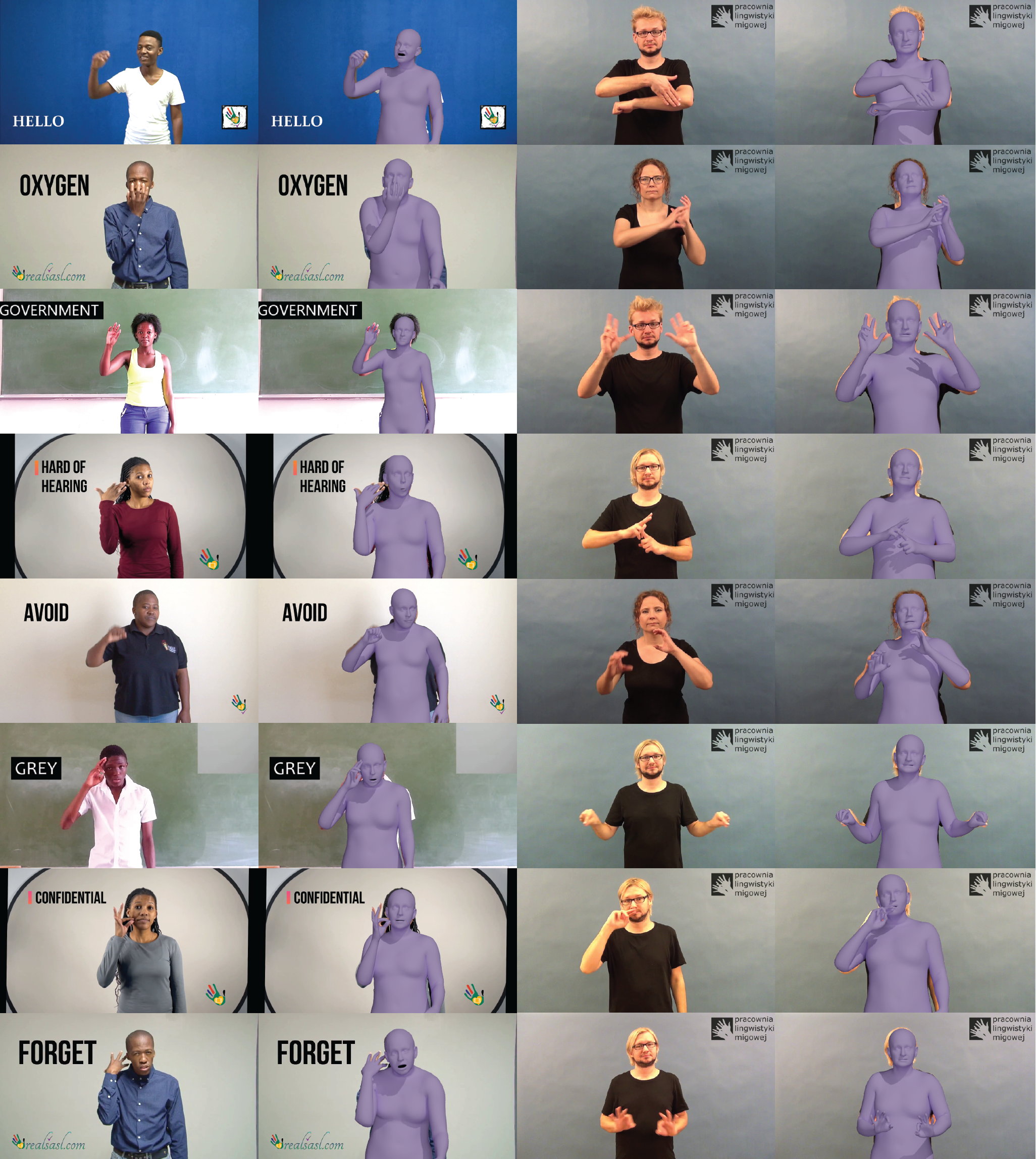}
    \end{minipage}
    \vspace{-0.5 em}
    \captionof{figure}{Additional examples on the Real SASL and CDPSL sign-language dictionaries.}
    \label{fig:examples1}
    \vspace*{+01.60em}
}]
\twocolumn[{
    \renewcommand\twocolumn[1][]{#1}
    \maketitle
    \centering
    \vspace{-0.2em}
    \begin{minipage}{1.00\textwidth}
    \centering
    \includegraphics[width=\linewidth]{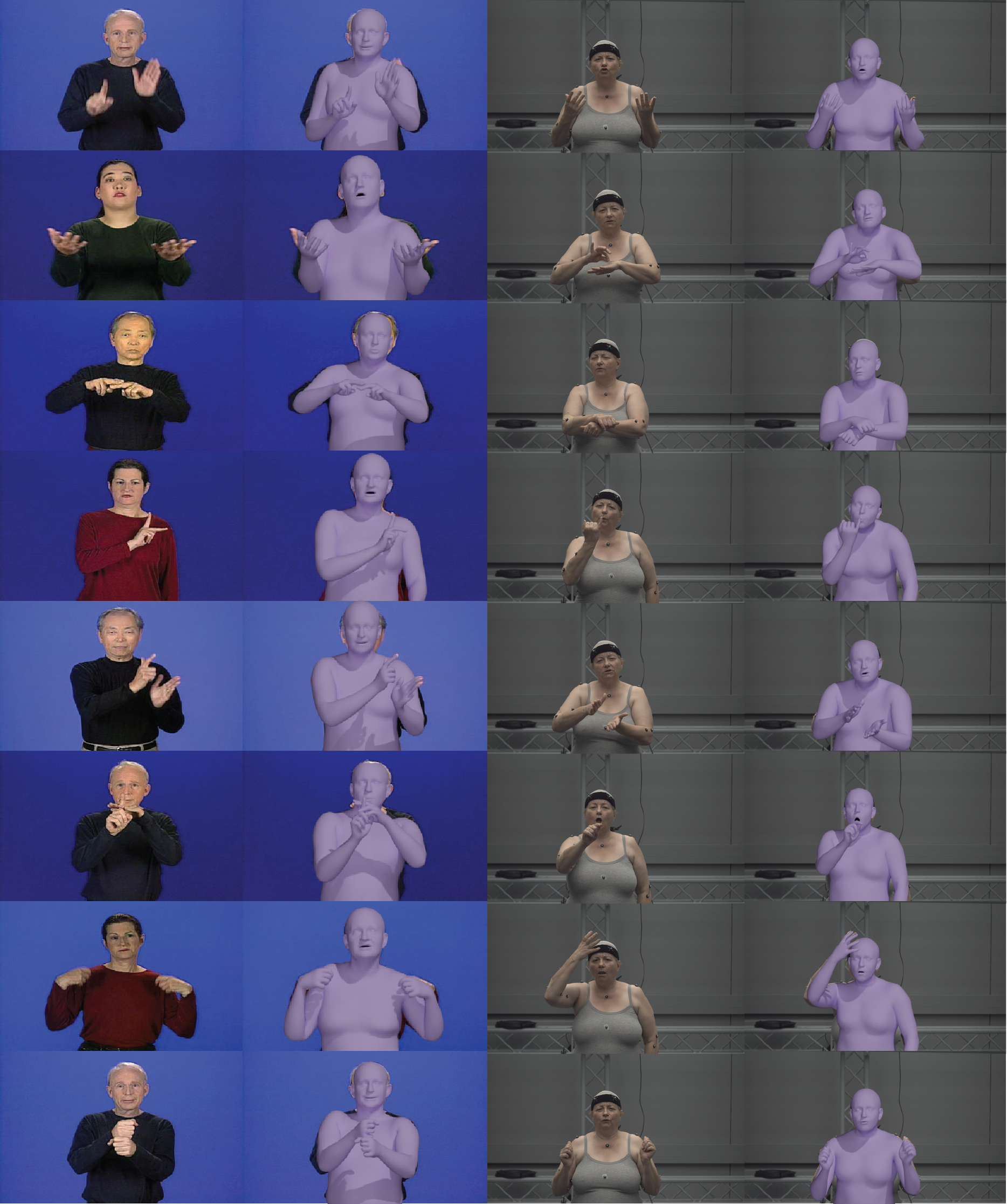}
    \end{minipage}
    \vspace{-0.1in}
    \captionof{figure}{Additional examples on The American Sign Language Handshape Dictionary and our captured dataset.}
    \label{fig:examples2}
    \vspace*{+01.50em}
}]

\end{document}